\documentclass[preprint,12pt]{elsarticle}

\usepackage{graphicx}
\usepackage{amsmath}
\usepackage{graphicx}
\usepackage{lipsum}
\usepackage{rotating}
\usepackage{hyperref}
\usepackage{xcolor}
\usepackage{lineno}
\usepackage{pifont}

\usepackage{array}
\usepackage{multirow}
\newcolumntype{?}{!{\vrule width 2pt}}

\begin{document}
	
	\title{Evaluating Pre-trained Convolutional Neural Networks and Foundation Models as Feature Extractors for Content-based Medical Image Retrieval}
	
	\author[1]{Amirreza Mahbod\corref{cor1}}
	\ead{amirreza.mahbod@dp-uni.ac.at}
	\cortext[cor1]{Corresponding author}
	\author[2]{Nematollah Saeidi}
	\author[3,4]{Sepideh Hatamikia}
	\author[1]{Ramona Woitek}

	\address[1]{Research Center for Medical Image Analysis and Artificial Intelligence, Department of Medicine, Faculty of Medicine and Dentistry, Danube Private University, Krems an der Donau, Austria}
	\address[2]{Artificial Intelligence Department, Faculty of Computer Engineering, University of Isfahan, Isfahan, Iran}
	\address[3]{Research Center for Clinical AI-Research in Omics and Medical Data Science, Department of Medicine, Faculty of Medicine and Dentistry, Danube Private University, Krems an der Donau, Austria}
	\address[4]{Austrian Center for Medical Innovation and Technology, Wiener Neustadt, Austria}
	
	\date{}
	
\begin{abstract}
Medical image retrieval refers to the task of finding similar images for given query images in a database, with applications such as diagnosis support. While traditional medical image retrieval relied on clinical metadata, content-based medical image retrieval (CBMIR) depends on image features, which can be extracted automatically or semi-automatically. Many approaches have been proposed for CBMIR, and among them, using pre-trained convolutional neural networks (CNNs) is a widely utilized approach. However, considering the recent advances in the development of foundation models for various computer vision tasks, their application for CBMIR can also be investigated.

In this study, we used several pre-trained feature extractors from well-known pre-trained CNNs and pre-trained foundation models and investigated the CBMIR performance on eight types of two-dimensional (2D) and three-dimensional (3D) medical images. Furthermore, we investigated the effect of image size on the CBMIR performance.

Our results show that, overall, for the 2D datasets, foundation models deliver superior performance by a large margin compared to CNNs, with the general-purpose self-supervised model for computational pathology (UNI) providing the best overall performance across all datasets and image sizes. For 3D datasets, CNNs and foundation models deliver more competitive performance, with contrastive learning from captions for histopathology model (CONCH) achieving the best overall performance. Moreover, our findings confirm that while using larger image sizes (especially for 2D datasets) yields slightly better performance, competitive CBMIR performance can still be achieved even with smaller image sizes. Our codes to reproduce the results are available at: \url{https://github.com/masih4/MedImageRetrieval}.

\end{abstract}
	
\maketitle
	
\section{Introduction}
Image retrieval refers to the process of finding similar images in a large database. Traditional image retrieval models were based on image metadata, but these approaches were often inefficient and required image metadata labeling, which could be very time-consuming~\cite{Vishraj2022}. Content-based image retrieval (CBIR), on the other hand, relies on the descriptive characteristics and features of images, which do not require specific labeling and can be handled in an unsupervised manner. CBIR can be applied in various domains, including satellite imagery, natural product search, surveillance, and social media~\cite{5580313}.

Content-based medical image retrieval (CBMIR) refers to the application of CBIR in the medical imaging domain. Compared to other medical computer vision tasks such as classification or segmentation, less research has been conducted on CBMIR. However, CBMIR models have a distinct advantage over classification models, for example. The output of a classification model for a given query image consists of probabilities for each class and, usually, through thresholding, provides a determined label for the test image. On the other hand, the output of a CBMIR model for a given query image consists of the top-$k$ relevant images ($k$ can be defined by the user, e.g., 3 or 5), usually retrieved from a large database of images. In many cases, the database also stores clinical metadata or diagnostic reports alongside the images, which can complement the model output as auxiliary information~\cite{10.1007/978-3-030-32239-7_61}. As a result, CBMIR provides medical experts and biologists with more comprehensive insights into diverse patterns and features, which allows them to identify the most similar patterns across different images. In contrast to classification models, which only produce labels for given images as black-box models, CBMIR keeps clinicians and medical experts in the loop for grading or classification, which is often necessary for artificial intelligence applications in healthcare. CBMIR models facilitate the diagnostic workflow by providing additional information; however, medical experts are still required to provide the final diagnostic conclusion rather than replacing medical experts with a machine learning or deep learning model~\cite{tang2020clinician, Herington1848}. Besides these applications, CBMIR can also be used for other purposes, such as image-based training of medical professionals and treatment planning~\cite{doi:10.1148/rg.253045071}. From a technical perspective, one advantage of CBMIR over classification or segmentation is that it is usually performed in an unsupervised manner, whereas classification and segmentation models typically require labeled data for training or fine-tuning.


A standard CBIR or CBMIR model typically consists of three main building blocks: an image pre-processing pipeline, feature extraction, and feature similarity measurements between query images and stored images in the database using various distance-based methods. While early models primarily relied on handcrafted features including color, texture, shape, and spatial information, most recent models use automatically extracted features from pre-trained models, such as pre-trained convolutional neural networks (CNNs) or vision transformers (ViTs), which are referred to as deep features. Although deep features suffer from low explainability and interpretability, they have demonstrated superior performance in many computer vision tasks, including image retrieval, due to their better generalization capability~\cite{9089643, 10.1145/2647868.2654948, 9933854}. Many pre-trained CNNs, such as ResNet family models, EfficientNet family models, or ViT family models, have been exploited as retrieval feature extractors in previous studies~\cite{Kalra2020, saeidi2024breast, denner2024leveraging}. In general, compared to CBIR, which has been widely studied, less research has been conducted on CBMIR, and the majority of recent CBMIR studies rely on CNNs as feature extractors~\cite{10635170, Hegde2019}.

Besides standard pre-trained CNNs and ViTs, which are typically trained on the ImageNet dataset~\cite{Deng2009} in a supervised manner, there has been a significant amount of research focused on foundation models recently, which can also extract deep features. A foundation model refers to a large-scale pre-trained model that can be used for various applications, typically in computer vision or natural language processing. Foundation models are usually trained in a self-supervised or unsupervised manner using approaches such as contrastive language-image pre-training (CLIP)~\cite{pmlr-v139-radford21a}, Mixup-CLIP~\cite{10692575} or knowledge distillation with no labels (DINO)~\cite{9709990, oquab2023dinov2}. Various foundation models have been developed or applied for tasks in medical computer vision, and some recent studies have explored their use in radiological image retrieval~\cite{denner2024leveraging, 10635170, khun2024content}. However, to our knowledge, no studies have thoroughly investigated the retrieval performance of both radiological and non-radiological medical images using multiple recently developed foundation models. This represents a significant gap in research, as understanding how different foundation models perform across diverse medical imaging domains is crucial for developing robust retrieval systems. In addition to this gap, another understudied aspect of medical image retrieval, particularly for foundation models, is the effect of image size on retrieval performance. Most foundation models were originally trained on $224\times224$ images, and the impact of using smaller images on their general performance has not been well-investigated.


In this study, we conducted a comprehensive analysis of CBMIR for various types of two-dimensional (2D) and three-dimensional (3D) radiological and non-radiological medical images, using both pre-trained CNNs (as benchmarks) and recently developed foundation models that have not been thoroughly investigated in previous studies. We evaluated their retrieval performance for zero-shot retrieval without any post-processing, fine-tuning, or retraining. Furthermore, we investigated the effect of image size (ranging from very small images with only 28 pixels in one dimension to standard 224-pixel images) on retrieval performance across eight types of 2D and 3D medical images.

In summary, the main contributions and findings of this study are as follows:
\begin{itemize}
\item We compared the CBMIR performance of well-known pre-trained CNNs and recently developed medical and non-medical foundation models.

\item We evaluated the impact of image size on the retrieval performance of both CNNs and foundation models.
\item We performed comprehensive experiments on a variety of radiological and non-radiological medical images in both 2D and 3D formats.
\item Our main findings indicate that for 2D datasets, foundation models achieved superior retrieval performance compared to benchmark pre-trained CNN models. However, for 3D datasets, foundation models and CNNs show comparable performance. Our experiments on image size also revealed that while competitive retrieval performance can still be achieved with small-sized images, increasing image size leads to slightly better performance, particularly for 2D datasets.
\item To ensure reproducibility and facilitate the application of our method to new datasets, we have made our implementation publicly available.

\end{itemize}

\section{Material and Methods}
The general workflow of the method is shown in Figure~\ref{fig:flowchart}, and the detailed descriptions are as follows:

\begin{figure}
	\centering
	\includegraphics[width=0.6\linewidth]{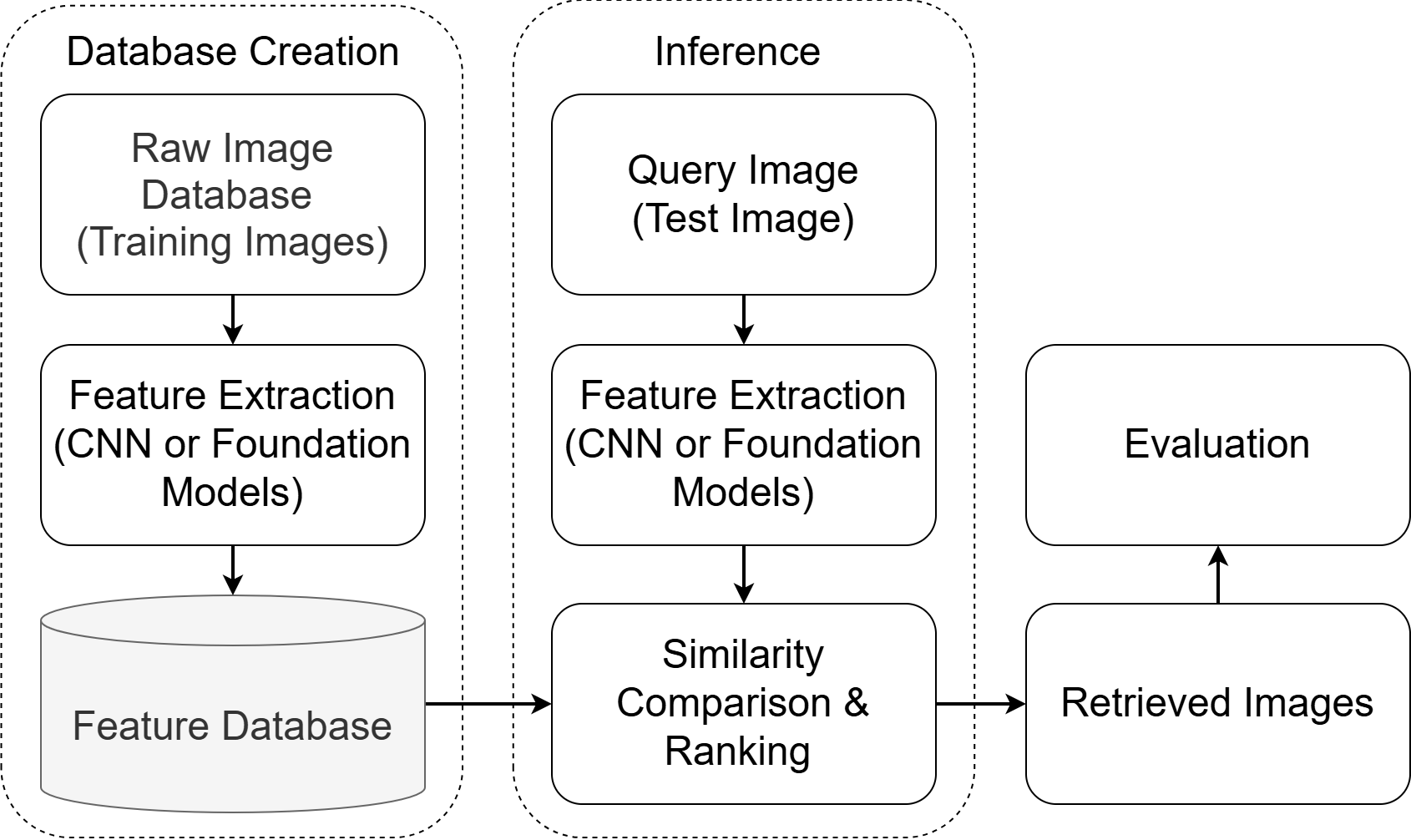}
	\caption{The general workflow of the image retrieval method. CNN: Convolutional Neural Networks. }
	\label{fig:flowchart}
\end{figure}

\subsection{Datasets}
\begin{figure}[t!]
	\centering
	\begin{tabular}{ccccc}
		 BreastMNIST & 	PneumoniaMNIST   &  	RetinaMNIST	 &  DermaMNIST	  \\
		\includegraphics[width=3cm]{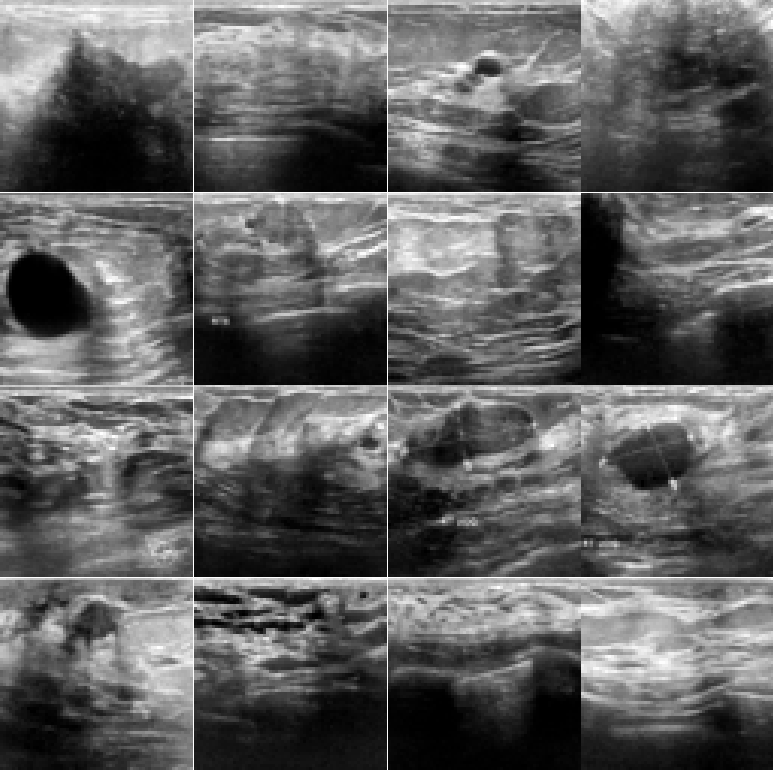} &
		\includegraphics[width=3cm]{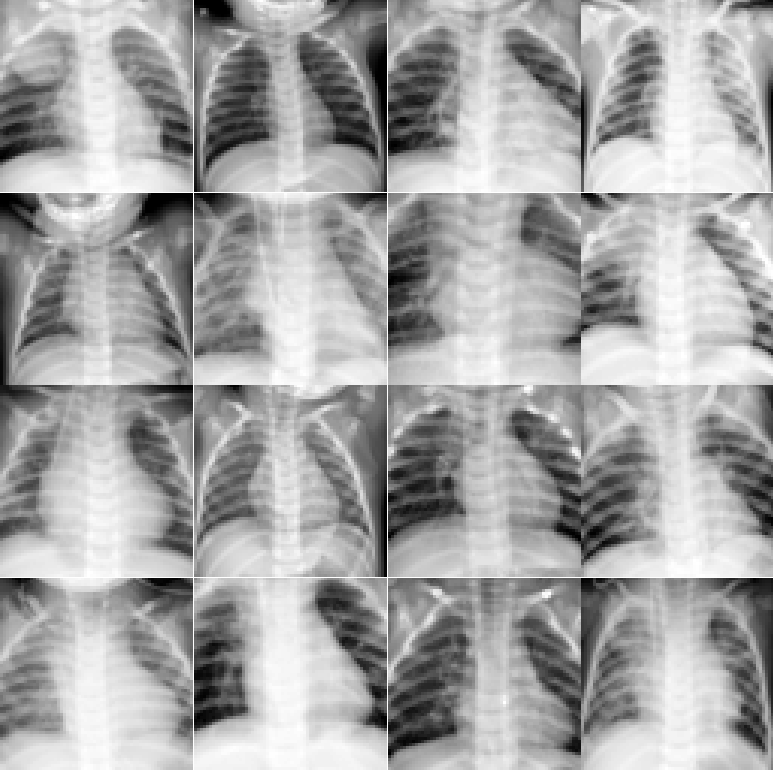} &
		\includegraphics[width=3cm]{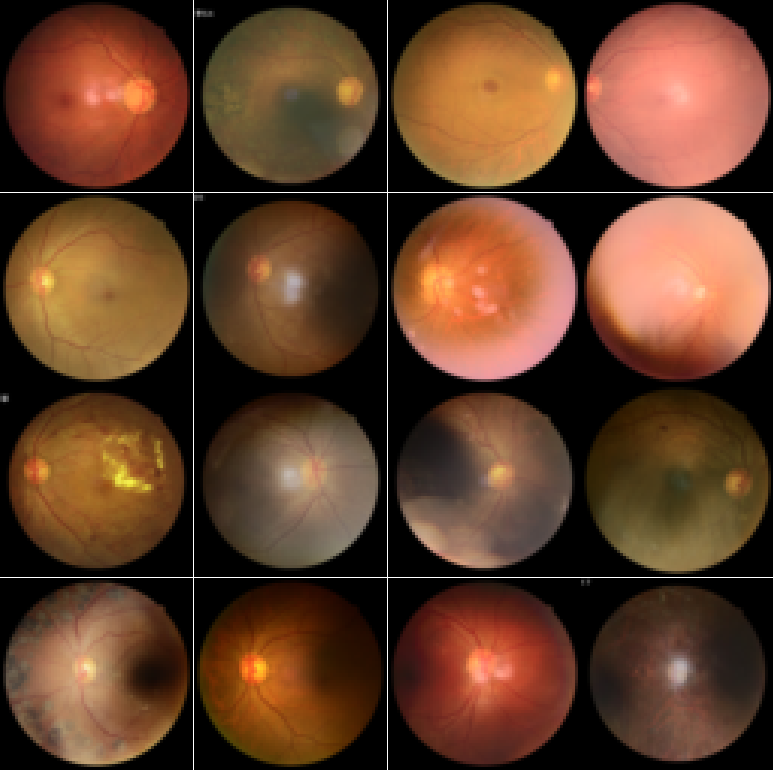} &
		\includegraphics[width=3cm]{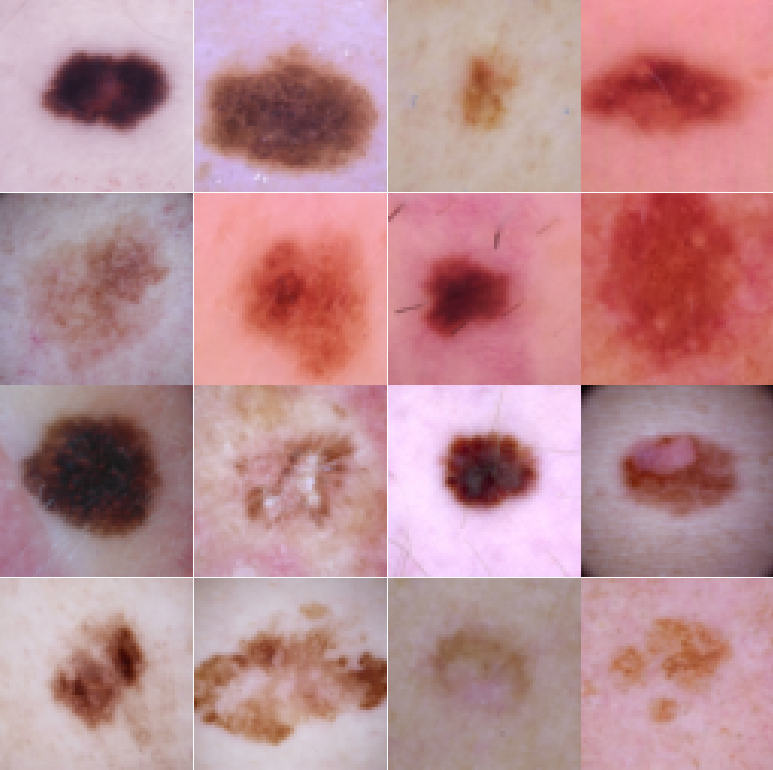}  \\
		\hline
		 BloodMNIST	   &  	PathMNIST &	 AdrenalMNIST3D & 	 SynapseMNIST3D \\
		\includegraphics[width=3cm]{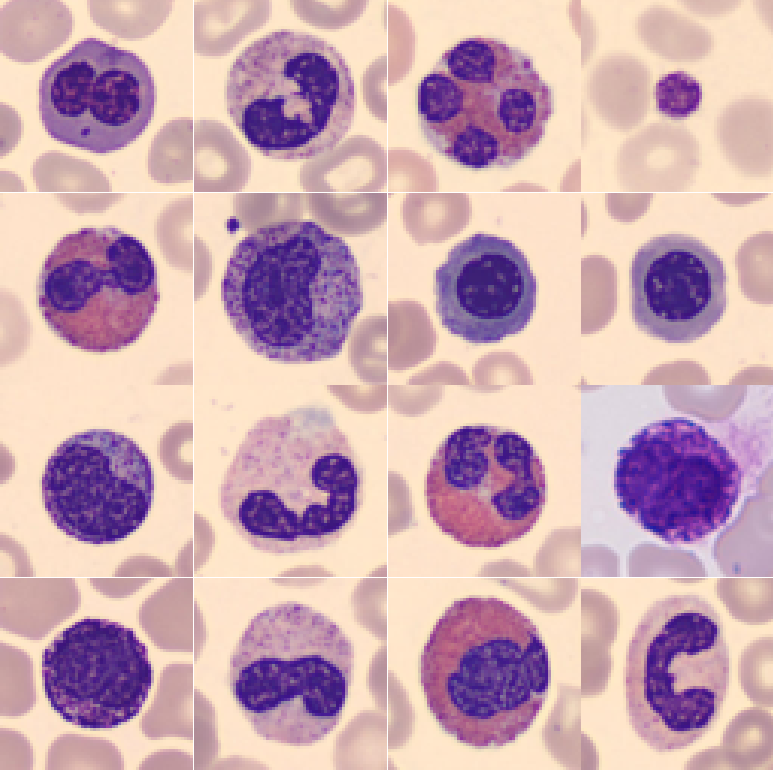} &
		\includegraphics[width=3cm]{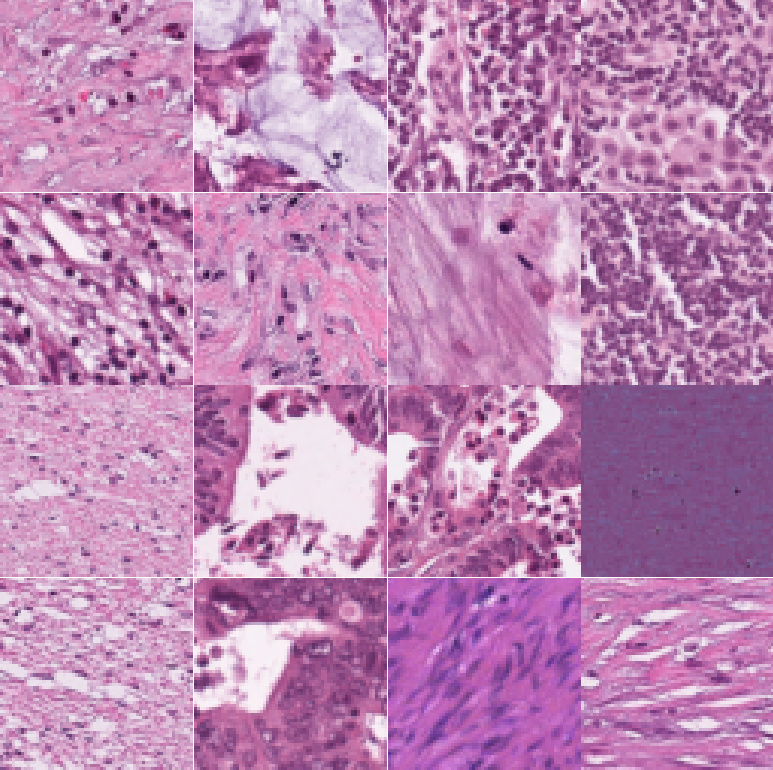} &
		\includegraphics[width=3cm]{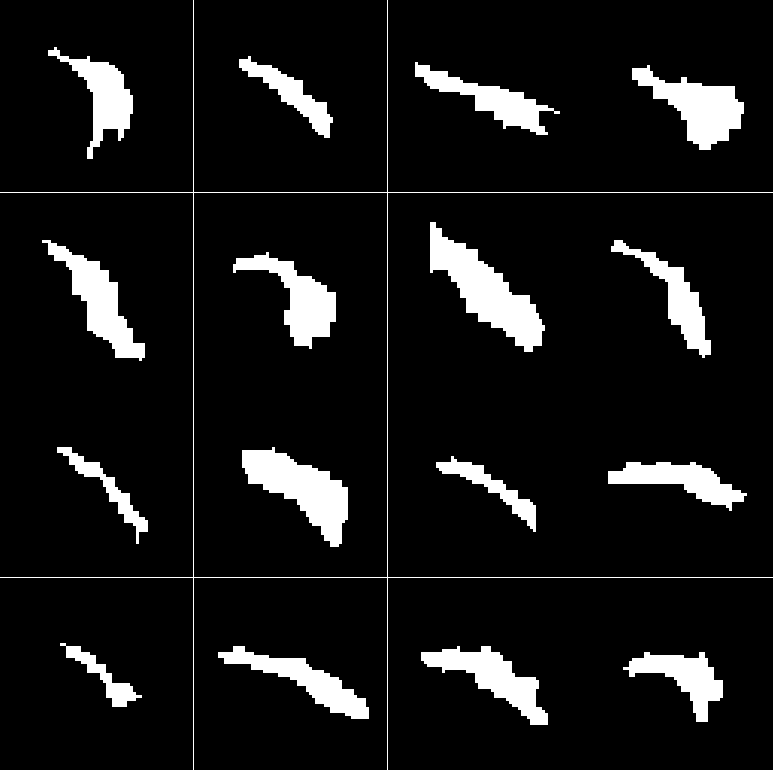} &
		\includegraphics[width=3cm]{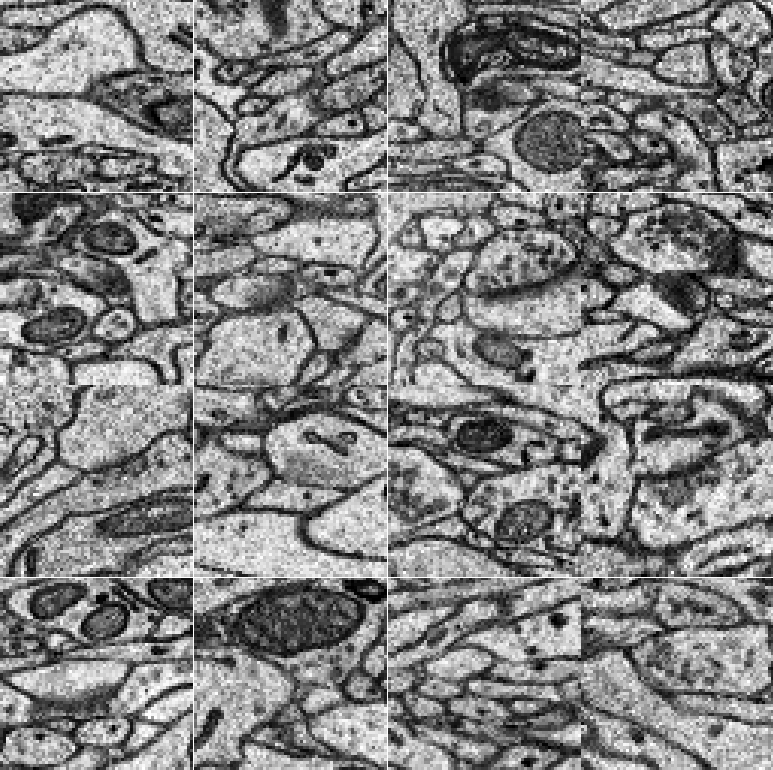}

	\end{tabular}
	\caption{Example images from the MedMNIST V2 dataset. For the last two images from the 3D datasets, middle slices from random samples were chosen.}
	\label{fig:examples}
\end{figure}

We used a subset of eight datasets introduced in the MedMNIST V2 study~\cite{Yang2023, 9434062}. We selected datasets from various imaging modalities, including colon pathology, dermatoscopy, chest X-ray, fundus imaging, breast ultrasound, blood cell microscopy, abdominal CT, and electron microscopy, to investigate image retrieval performance across a wide range of medical images, both in 2D (six datasets) and 3D (two datasets). The basic statistics of these datasets, including the imaging type (2D or 3D), number of training/testing images per dataset, number of classes, and imaging modality are listed in Table~\ref{tab:datasets}, and example images from each dataset are shown in Figure~\ref{fig:examples}. In the latest version of MedMNIST V2, 2D datasets are provided in various sizes of 28$\times$28, 64$\times$64, 128$\times$128, and 224$\times$224 pixels, and 3D datasets are provided in two sizes of 28$\times$28$\times$28 and 64$\times$64$\times$64 pixels. Further details about each sub-dataset can be found in the respective publications as referenced in Table~\ref{tab:datasets}. It should be noted that we evaluated the performance of the retrieval model on the test images, and the training images were used to build the database for image search. Moreover, we used the original training and testing splits of the MedMNIST V2 datasets in this study.

\begin{table*}[]
	\caption[]{Selected datasets from the MedMNIST V2~\cite{Yang2023}. In the dataset names in the first column, "MNIST" is removed (e.g., Breast refers to BreastMNIST, and Adrenal3D refers to AdrenalMNIST3D). E. Microscopy = Electron Microscopy. }
	\label{tab:datasets}
	\begin{tabular}{lccccccc}
		\hline
		\textbf{Dataset} & \textbf{Type} & \textbf{\# Train/Test} & \textbf{\# Classes} & \textbf{Modality}   \\
		\hline
		Breast~\cite{ALDHABYANI2020104863}         & 2D  & 546/156      & 2  & Ultrasound          \\
		Pneumonia~\cite{KERMANY20181122}           & 2D  & 4,708/624    & 2  & X-Ray               \\
		Retina~\cite{LIU2022100512}                & 2D  & 1,080/400    & 5  & Fundus
		imaging      \\
		Derma~\cite{tschandl2018ham10000}          & 2D  & 7,007/2,005  & 7  & Dermatoscopy        \\
		Blood~\cite{ACEVEDO2020105474}             & 2D  & 11,959/3,421 & 8  & Microscopy          \\
		Path~\cite{10.1371/journal.pmed.1002730}   & 2D  & 89,996/7,180 & 9  & Pathology           \\
		\hline
		Adrenal3D~\cite{Yang2023}                  & 3D  & 1,188/298    & 2  & CT                  \\
		Synapse3D~\cite{Yang2023}                  & 3D  & 1,230/352    & 2  & E. Microscopy \\

		\hline
	\end{tabular}
\end{table*}

\subsection{Pre-trained models as feature extractors}
\label{sec:pretrained}
We used both well-known pre-trained CNN models (as benchmarks) and a number of foundation models as feature extractors to compare their retrieval performance on the selected sub-datasets from the MedMNIST V2. The pre-trained CNN models include VGG19~\cite{Simonyan2014}, ResNet50~\cite{He2016}, DenseNet121~\cite{huang2017densely}, and EfficientNetV2M~\cite{Tan2021}, while the pre-trained foundation models comprise MedCLIP~\cite{wang2022medclip}, BioMedCLIP~\cite{zhang2023large}, OpenCLIP~\cite{ilharco_gabriel_2021_5143773, 10205297}, CONCH~\cite{Lu2024}, and UNI~\cite{Chen2024}. For foundation model selection, we chose models that were not thoroughly investigated in previous studies for various types of medical image retrieval. We excluded some foundation models such as DINO V1~\cite{9709990}, DINO V2~\cite{oquab2023dinov2}, SAM~\cite{Kirillov_2023_ICCV}, MEDSAM~\cite{Ma2024}, Virchow~\cite{Vorontsov2024}, and MAE~\cite{9879206}, as well as pre-trained ViT-based models like the original ViT~\cite{dosovitskiy2020image} or Swin Transformers~\cite{9710580}, because prior studies have demonstrated their inferior performance in medical image retrieval compared to the models selected for this study~\cite{10635170, Chen2024, denner2024leveraging}.
The basic statistics of the selected CNN and foundation models are listed in Table~\ref{tab:models}, and a brief description of each is provided in the following.

\begin{table*}[]
	\caption[]{Basic statistics of the utilized convolutional neural networks (upper part) and foundation models (lower part) in this study. Nat. = Natural, Med. = Medical.
	}
	\label{datasets}
	\begin{tabular}{lllccccc}
		\hline
		\textbf{Feature extractor} & \textbf{\begin{tabular}[c]{@{}l@{}}Pre-train \\ domain\end{tabular}} & \textbf{Type} & \textbf{Feature size}   \\
		\hline
		VGG19~\cite{Simonyan2014}                              & Nat.            & Vision               &512 \\
		ResNet50~\cite{He2016}                                 & Nat.            & Vision               &2048 \\
		DenseNet121~\cite{huang2017densely}                    & Nat.            & Vision               &1024 \\
		EfficientNetV2M~\cite{Tan2021}                         & Nat.            & Vision               &1280 \\ \hline
		MedCLIP~\cite{wang2022medclip}                         & Med.(radiology) & Vision/Language      &512    \\
		BioMedCLIP~\cite{zhang2023large}                       & Med. (various)      & Vision/Language      &512 \\
		OpenCLIP~\cite{ilharco_gabriel_2021_5143773, 10205297} & Nat.            & Vision/Language      &1024 \\
		CONCH~\cite{Lu2024}                                   & Med.(pathology) & Vision/Language      &512    \\
		UNI~\cite{Chen2024}                                    & Med.(pathology) & Vision               &1024 \\

		\hline
	\end{tabular}
		\label{tab:models}
\end{table*}

\subsubsection{VGG19}
Visual Geometry Group or VGG is one of the most well-known CNN-based models that has been widely used for various computer vision applications, including content-based image retrieval~\cite{ALZUBI201795}. The original VGG model was mainly designed for the image classification task and contains convolutional layers, max-pooling layers, and fully connected layers. Depending on the depth of the model, various variants became available, and in this study, we used one of its most utilized variants, VGG19. To use it as a feature extractor, we employed the pre-trained model based on the ImageNet dataset (1.2 million labeled training images with 1000 classes)~\cite{Deng2009} and extracted the features from the last convolutional layer. We used average pooling to flatten the features from an 8$\times$8$\times$512 feature dimension to 512 features, as stated in Table~\ref{tab:models}.

\subsubsection{ResNet50}
Residual networks or ResNets are also widely used CNN-based models and they have also been used for CBMIR in previous studies~\cite{Agrawal2022, mahbod2018breast}. Besides the main VGG building blocks, they are equipped with skip connections and identity mapping to better distribute the gradient in the network, especially for very deep models. Like VGG, depending on the depth, ResNet models are available in various variants, and we used the ResNet50 model in this study. We extracted the features from the pre-trained ResNet50 (pre-trained on the ImageNet dataset) from the last convolutional layer and flattened the feature map from 7$\times$7$\times$2048 to a 2048 feature vector using the average pooling operation.

\subsubsection{DenseNet121}
Densely connected convolutional networks or DenseNets are another widely used CNN-based model with various applications such as medical image classification and retrieval~\cite{9313211, 9412307}. DenseNets extend the skip connections of the ResNet model for better gradient flow by connecting each layer to every other layer in the model using feed-forward connections. In this study, we used DenseNet121, one of the most widely employed variants of the DenseNet model family. For image feature extraction, we used the feature map before the fully connected layer of the DenseNet121. Similar to the previous two models, we used average pooling to flatten the feature map from 8$\times$8$\times$1024 to 1024 features.

\subsubsection{EfficientNetV2M}
EfficientNet is a family of CNN models initially designed to systematically and uniformly scale all dimensions of the model (depth, width, and resolution) using compound coefficients~\cite{tan2019efficientnet}. In the original study, neural architecture search~\cite{zoph2016neural} was employed to find a baseline model (EfficientNetB0) consisting of convolutional and mobile inverted bottleneck (MB) convolutional blocks~\cite{Ekoputris2018}, which was then scaled up to larger models for better performance (EfficientNetB1-B7). In the second version of the EfficientNet family (EfficientNetV2), a combination of Fused-MB convolutional blocks~\cite{gupta2020accelerator} (in the initial stages) and MB convolutional blocks (in the later stages) was employed and the network was optimized for improving training speed, parameter efficiency and performance. Other techniques, such as non-uniform scaling strategy, progressive learning (gradually increasing the image resolution during training), and exploiting regularizers (dropout, RandAugment, mixup), were also utilized to further boost the performance. In this study, we utilized EfficientNetV2M, which has been shown to deliver excellent performance for e.g. image classification and retrieval~\cite{Tan2021, 10.1007/978-3-031-20233-9_45} while being computationally efficient. Like other CNN-based models, we used average pooling to convert the feature maps from 8$\times$8$\times$1280 to 1280 features.

\subsubsection{MedCLIP}
Vision-language models such as CLIP~\cite{pmlr-v139-radford21a} have shown excellent performance for a variety of computer vision tasks. CLIP consists of two main building blocks, a vision encoder and a text encoder that map the input images and texts to a common latent space. Using contrastive learning, the model is trained to maximize the similarity between paired images and texts. The original CLIP model was trained on 400 million general image-text pairs (named as WIT-400M) extracted from the internet. However, in the context of medical data, finding large image-text paired datasets is very challenging. In contrast to CLIP, MedCLIP is trained with three types of datasets, including medical image-text, only medical image, and only medical text datasets. To train the model, two similarity matrices were created: one referred to as the semantic similarity matrix and the other as the predicted similarity matrix. To build the semantic similarity matrix, the MetaMap~\cite{10.1136/jamia.2009.002733} approach was used to extract entities from texts and images in all three types of datasets. To build the predicted similarity matrix, a text encoder (BioClinicalBERT\footnote{\url{https://huggingface.co/emilyalsentzer/Bio_ClinicalBERT}}) and a vision encoder (Swin Transformer~\cite{9710580}) were used. The utilized image encoder was pre-trained on the ImageNet dataset. Finally, to train the model, semantic matching loss was used instead of information noise-contrastive estimation (InfoNCE) loss used in the CLIP study. The model was pre-trained using two X-ray datasets, namely, MIMIC-CXR (around 600K image-text pairs) and CheXpert (around 223K images). To extract features using the pre-trained MedCLIP model, we used the vision encoder embedding, which yields 512 features for each input image.

\subsubsection{BioMedCLIP}
This model is an extension of the CLIP study, specifically designed for medical data. BioMedCLIP pre-training was performed using the PMC-15M dataset, which contains around 15 million image-text pairs automatically extracted from 4.4 million PubMed articles. BioMedCLIP employs a vision transformer (ViT-B/16-224)~\cite{dosovitskiy2020image} as the vision encoder and PubMedBERT~\cite{10.1145/3458754} with an extended token size of 256 as the text encoder. To extract image features from BioMedCLIP, we used the vision encoder, which yields 512 features for each input image.

\subsubsection{OpenCLIP}
Previous studies have shown that using the original CLIP model for downstream medical applications is suboptimal~\cite{wang2022medclip, zhang2023large, Lu2024}. However, it should be noted that CLIP pre-training was performed on the WIT-400M dataset. In a recently published study~\cite{10205297}, the effects of model scaling, data scaling, and training duration (number of samples seen during training) on CLIP's performance for various downstream tasks were investigated. In this study, we chose one of the best configurations based on the reported ImageNet zero-shot classification accuracy. Specifically, we used the pre-trained model available on the OpenCLIP repository~\cite{ilharco_gabriel_2021_5143773}, where pre-training was performed on the LAION-2B dataset~\cite{schuhmann2022laionb} with the ViT-G/14-224 vision encoder~\cite{9880094} and 34 billion samples seen during training. The extracted features for each image with this pre-trained model had a size of 1024.

\subsubsection{CONCH}
Contrastive learning from captions for histopathology or CONCH is a vision-language foundation model trained on over 1.17 million histological image-text pairs sourced from in-house educational notes and the PubMed central open access dataset. Similar to pre-trained CNNs, which are trained on natural images and used for various applications such as medical image analysis, we also considered using CONCH and UNI (see next section) as feature extractors for a variety of medical image types. The dataset creation to train the CONCH model involved three main steps. First, using a
YOLO V5 model~\cite{7780460}, histological images were detected. Then a GPT-style model was utilized to split the image captions~\cite{10.1093/bib/bbac409}, and finally, a CLIP model was employed to match the images with their corresponding captions. The model training was conducted using the CoCa approach~\cite{yu2022coca}, which includes a vision encoder (ViT-B/16), a GPT-style text encoder, and a GPT-style text decoder. Unlike CLIP, which only uses contrastive loss to maximize cosine similarity scores for pairing images and text, CoCa is also equipped with a captioning objective to generate proper text for input images. The vision encoder of the model was pre-trained on 16 million in-house images using the iBOT self-supervised approach~\cite{zhou2021ibot}, while the language model was pre-trained using more than 550,000 surgical histopathology reports from Massachusetts General Hospital and over 400,000 PubMed abstracts related to histopathology. For our study, we used the vision encoder of the CONCH model to generate 512 features for the input images.

\subsubsection{UNI}
General-purpose self-supervised model for computational pathology or UNI is a large-scale visual pre-trained model, originally trained on histological images. UNI was trained using the Mass-100K dataset, which contains more than 100 million H\&E-stained image patches extracted from over 100,000 whole slide images. The network training was based on the DINO V2~\cite{oquab2023dinov2} model, a self-supervised, student-teacher-based approach for pre-training. For a given image, we extracted features using the ViT-G/14 encoder of the UNI model, which yielded a feature vector of size 1024.

\subsection{Distance measurement}
As mentioned in Section~\ref{sec:pretrained}, we used pre-trained CNNs and foundation models for feature extraction from both 2D and 3D datasets and all these models were originally trained on 2D images.  For 2D datasets, we simply used the 2D training and testing images for feature extraction. For 3D datasets, however, we extracted features from each slice of a 3D train/test volume and concatenated these features to obtain the feature map for the entire 3D volume. While this approach can be computationally expensive, we selected it because our initial results indicated that applying feature reduction techniques such as principal component analysis (PCA)~\cite{MACKIEWICZ1993303}, autoencoders~\cite{WANG2016232}, t-Distributed Stochastic Neighbor Embedding (t-SNE)~\cite{JMLR:v9:vandermaaten08a}, or uniform manifold approximation and projection (UMAP)~\cite{McInnes2018} led to a decrease in performance (refer to Table S5 and Table S6 in the supplementary materials). To measure the distance between the extracted features of the query images/volumes and the stored features in the database, we used the cosine similarity index, as suggested by previous studies on medical image retrieval~\cite{denner2024leveraging, pmlr-v139-radford21a}. After calculating the index for all images (query samples and all samples in the training set), we ranked them, and the top-$k$ most similar images were selected as the output of the retrieval model.

\subsection{Evaluation}
While CBMIR was completely performed in an unsupervised manner, the image class labels of the query and database images were used only for evaluation. If the output of the retrieval model had an identical label to the query image label, then the CBMIR model was considered to have performed correctly. However, if the labels were mismatched, the output of the model was deemed incorrect. To evaluate the performance of the retrieval results quantitatively, we used well-known retrieval evaluation metrics, including top-\textit{k} mean average precision (mAP@\textit{k}), majority vote at the top \textit{k} search results (mMV@\textit{k}), and top-\textit{k} accuracy (ACC@\textit{k}), as suggested in previous studies~\cite{10.1007/978-3-030-32239-7_61, Kalra2020, saeidi2024breast, WANG2023102645}. The detailed formulas for these metrics are provided below.
	
mAP represents the average precision values calculated at the ranks of relevant images in the retrieved results. It is a widely used metric for assessing the effectiveness of a retrieval system. The formula for mAP is as follows:

\begin{equation}
	mAP@ k = \frac{1}{|Q|} \sum_{q=1}^{|Q|} \frac{1}{|R_q|} \sum_{k=1}^{|R_q|} \text{Precision}(k) \cdot \text{rel}(k)
\end{equation}

\noindent where $|Q|$ represents the number of queries, $R_q$ denotes the number of relevant images for query image  $q$ and $\text{Precision}(k)$ is the precision at rank $k$ (i.e., the ratio of relevant images retrieved within the top $k$ results to the total number of retrieved images up to rank $k$). Additionally, $\text{rel}(k)$ is an indicator function that equals 1 if the image at rank $k$ is relevant and 0 otherwise. The notation $|\cdot|$ indicates the length of a set.

mMV measures how often the primary diagnosis among the top-$k$ results matches the queried label, as defined below:

\begin{equation}
mMV@ k = \frac{1}{|Q|} \sum_{q=1}^{|Q|} 1\{L_q \in MV(\text{rel}(k))\} \quad 
\end{equation}

\noindent where $L_q$ represents the ground truth label of the image, and $MV$ indicates the predicted diagnosis obtained from the majority vote of the top-$k$ samples. Similar to mAP, $\text{rel}(k)$ is an indicator function that and $|\cdot|$ denotes the length of the set.

And finally top-\textit{k} accuracy can be defined as below:

\begin{equation}
	\operatorname{ACC}@ k=\frac{1}{|Q|} \sum_{q=1}^{|Q|} 1\left(L_q, \operatorname{TOP}\left(\operatorname{ans}_q[: k]\right)\right)
\end{equation}

where $\operatorname{TOP}\left(\operatorname{ans}_q\left[\begin{array}{ll}:&k]\end{array}\right)\right.$  returns top-$k$ retrieved results, and $|Q|$ and $L_q$ remain the same as defined above.

In our experiments, we used \textit{k} = 5 for mAP and mMV, and \textit{k} = 1, 3, 5 for ACC. As our initial results showed that ACC@1 is the strictest evaluation metric, we used it to report specific results.

\subsection{Implementation Details}
We used Keras/TensorFlow and PyTorch for CNN models and foundation models, respectively. For feature extraction from pre-trained CNNs, we used \textit{tf.Keras.applications}, while for foundation models, we utilized repositories introduced in the corresponding studies on GitHub~\footnote{\url{https://github.com/mlfoundations/open_clip},\\\url{https://github.com/mahmoodlab/UNI}, \\\url{https://github.com/RyanWangZf/MedCLIP}} or Hugging Face~\footnote{\url{https://huggingface.co/microsoft/BiomedCLIP-PubMedBERT_256-vit_base_patch16_224}}. 
Besides the TensorFlow implementation for CNN models for 2D and 3D datasets, we have also provided the corresponding PyTorch codes in our GitHub repository. However, the results reported for CNN models in this study are based on the TensorFlow implementation.

Regarding the datasets, we used the original training/test split of the MedMNIST V2 dataset for all experiments. The images for foundation models were resized to the required input size of the models (i.e., 224$\times$224 pixels), while for CNN models, the input images were used in their original sizes, with the exception of 28$\times$28 or 28$\times$28$\times$28 pixel images, which were resized to 32$\times$32 or 32$\times$32$\times$32 pixels as the minimum acceptable input size for pre-trained CNN models.

All experiments were conducted using a single workstation with an Intel Core i7-8700 3.20 GHz CPU, 32 GB of RAM, and a TITAN V Nvidia GPU card with 12 GB of memory. The code developed to generate and reproduce these results is available in our GitHub repository: \url{https://github.com/masih4/MedImageRetrieval}.

\section{Results \& Discussion}
\label{sec:Results}
We report the detailed retrieval performance based on image sizes and sub-datasets for all evaluation indices in Tables S1, S2, S3, and S4 in the supplementary materials available in our GitHub repository. The average results across all image sizes for 2D and 3D datasets are reported in Tables~\ref{tab:2d} and Tables~\ref{tab:3d}, respectively. We also report the ACC@1 results for each sub-dataset based on the utilized models and image sizes in Figures~\ref{fig:BreastMNIST} to Figures~\ref{fig:SynapseMNIST3D}. 
 
In the following subsections, we first provide a detailed discussion of the 2D results, 3D results, and merged results in Subsections~\ref{sec:2D Results}, \ref{sec:3D Results}, and \ref{sec:Merged Results}. Then, we present an overall discussion on the advantages and disadvantages of CNNs and foundation models used for the medical image retrieval task in Subsection~\ref{sec:overall}.

\subsection{2D Results}
\label{sec:2D Results}
\begin{figure}[]
	\centering
	\begin{tabular}{c}
		
		\includegraphics[width=\linewidth]{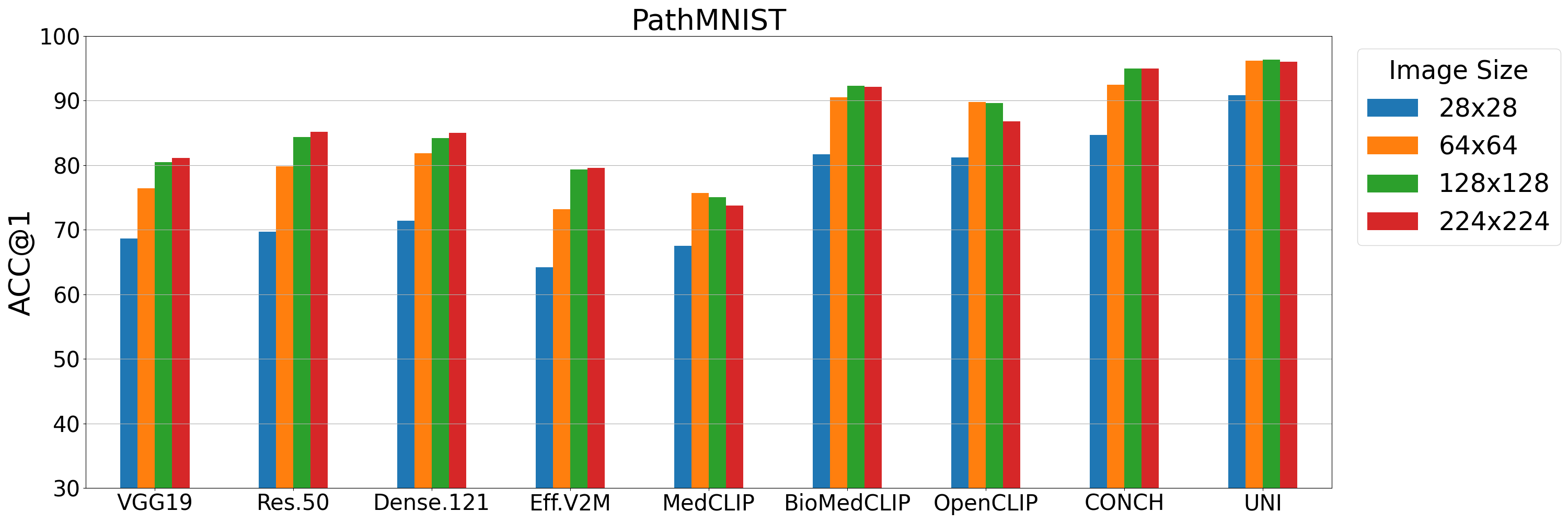} \\
	\end{tabular}
	\caption{ACC@1 performance based on different image sizes for the PathMNIST dataset. Res.50 = ResNet50, Dense.121 = DenseNet 121, Eff.V2M = EfficientNetV2M.}
	\label{fig:PathMNIST}
\end{figure}

For 2D results in Table~\ref{tab:2d}, we observe that all foundation models (except MedCLIP) deliver superior performance across all evaluation indices compared to pre-trained CNN models. Although MedCLIP is categorized as a foundation model, it is important to note that its training data size was significantly smaller compared to other foundation models. However, as shown in Tables S1 to S4 in the supplementary material and Figure~\ref{fig:PneumoniaMNIST}, MedCLIP interestingly delivers one of the best overall performances for the PneumoniaMNIST sub-dataset. The most likely reason is that the domain of MedCLIP’s pre-training data and the PneumoniaMNIST sub-dataset is identical (X-ray). According to the results in Table~\ref{tab:2d}, among the CNN models, DenseNet121 achieves the best retrieval performance, while EfficientNetV2M performs the worst. Among the foundation models, UNI achieves the highest retrieval performance, while MedCLIP performs the worst. Based on the mAP@5, mMV@5, ACC@1, ACC@3 and ACC@5 scores, the overall best foundation model (UNI) outperforms the best CNN model (DenseNet121) by 4.40\%, 3.65\%, 5.38\%, 2.29\%, 1.28\%. The superior performance of UNI is particularly interesting as it was only trained on histological images, yet it generalized well to other types of medical images, even outperforming BioMedCLIP, which was trained on various types of medical images. This finding confirms that medical foundation models trained on specific imaging modalities can generalize to others. Moreover, for histological images, as shown in Figure~\ref{fig:PathMNIST}, UNI and CONCH (both trained exclusively on histological images) deliver the best ACC@1 performance for the PathMNIST dataset across all sizes.
It is worth mentioning that, as seen in Table~\ref{tab:2d}, while UNI delivers the best overall performance (best mAP@5, best mMV@5, and best ACC@1), all other foundation models (except MedCLIP) perform only slightly worse. 

\begin{figure}[]
	\centering
	\begin{tabular}{c}
		
		\includegraphics[width=\linewidth]{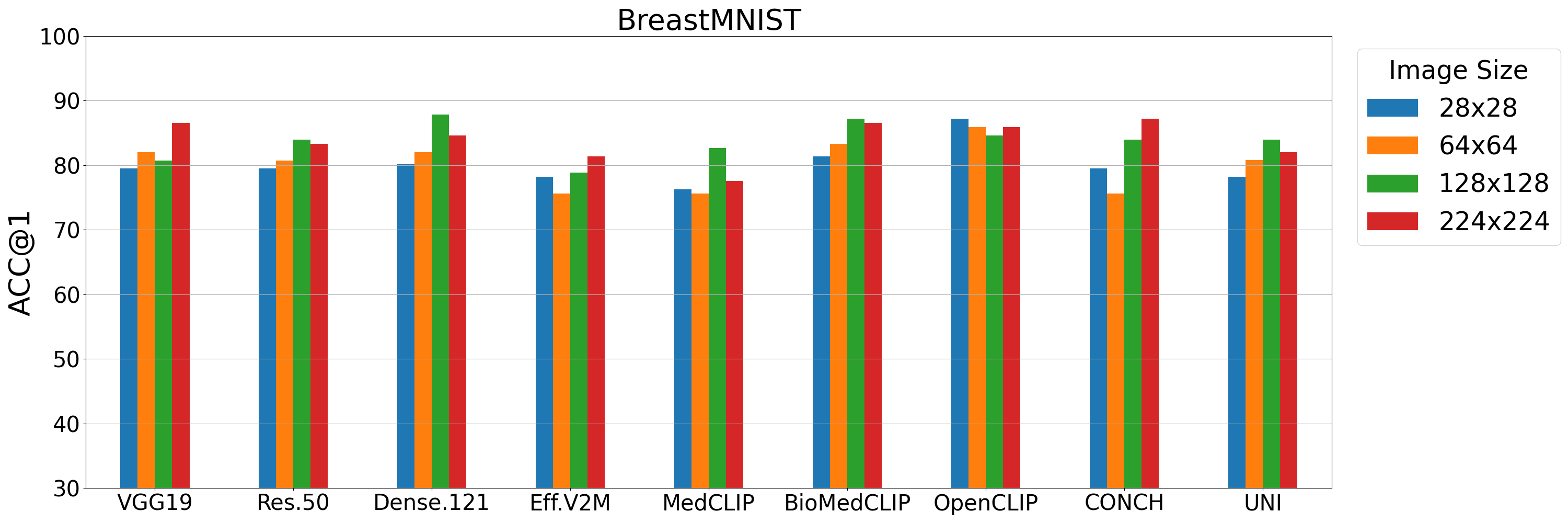} \\
	\end{tabular}
	\caption{ACC@1 performance based on different image sizes for the BreastMNIST dataset. Res.50 = ResNet50, Dense.121 = DenseNet 121, Eff.V2M = EfficientNetV2M. }
	\label{fig:BreastMNIST}
\end{figure}

\begin{figure}[]
	\centering
	\begin{tabular}{c}
		
		\includegraphics[width=\linewidth]{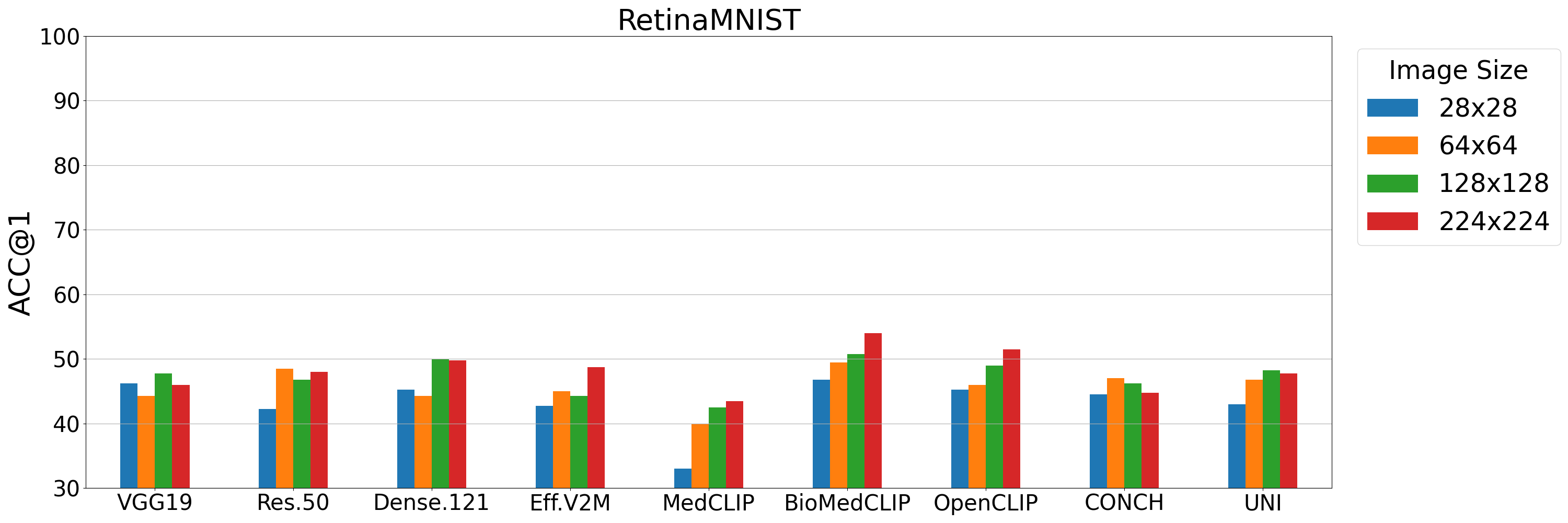} \\
	\end{tabular}
	\caption{ACC@1 performance based on different image sizes for the RetinaMNIST dataset. Res.50 = ResNet50, Dense.121 = DenseNet 121, Eff.V2M = EfficientNetV2M.}
	\label{fig:RetinaMNIST}
\end{figure}

\begin{figure}[h!]
	\centering
	\begin{tabular}{c}
		
		\includegraphics[width=\linewidth]{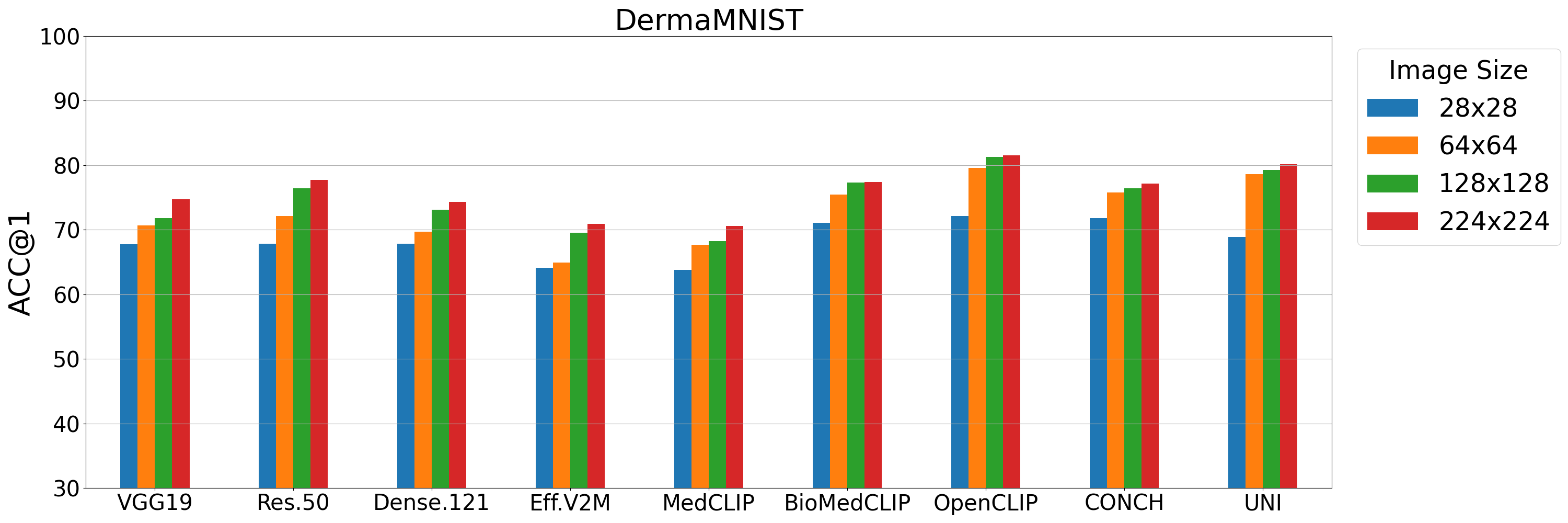} \\
	\end{tabular}
	\caption{ACC@1 performance based on different image sizes for the DermaMNISTT dataset. Res.50 = ResNet50, Dense.121 = DenseNet 121, Eff.V2M = EfficientNetV2M.}
	\label{fig:DermaMNIST}
\end{figure}

\begin{figure}[]
	\centering
	\begin{tabular}{c}
		
		\includegraphics[width=\linewidth]{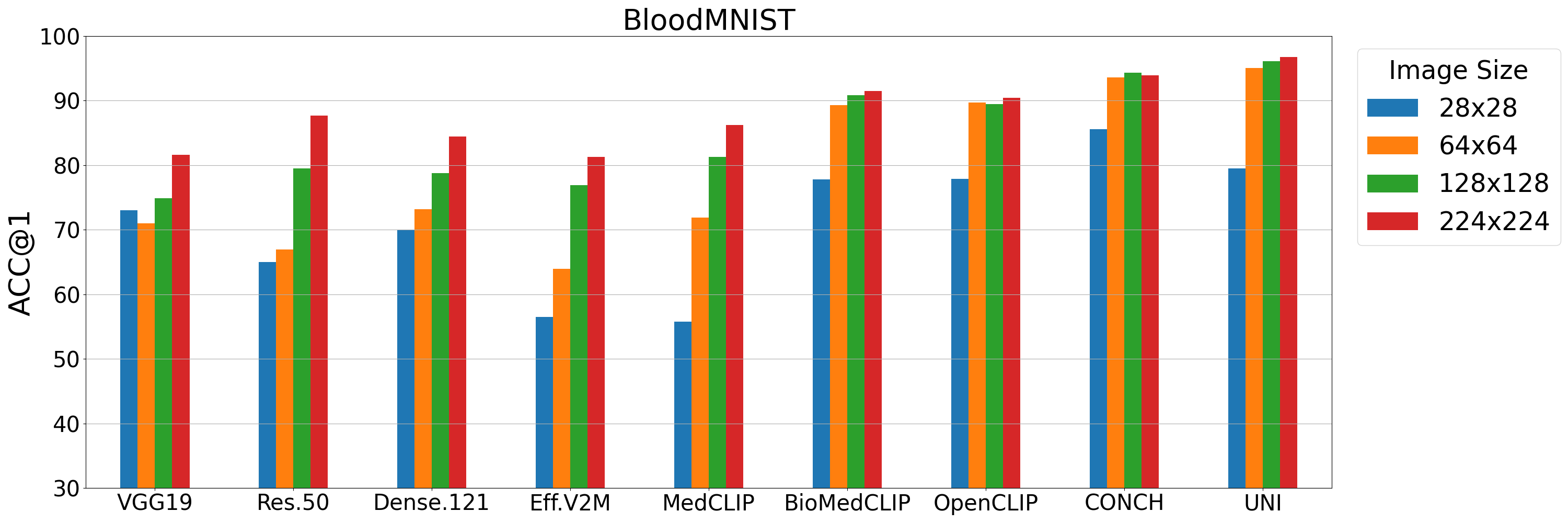} \\
	\end{tabular}
	\caption{ACC@1 performance based on different image sizes for the BloodMNIST dataset. Res.50 = ResNet50, Dense.121 = DenseNet 121, Eff.V2M = EfficientNetV2M.}
	\label{fig:BloodMNIST}
\end{figure}

The effect of image size on the performance of pre-trained CNNs for a number of computer vision tasks, such as classification, has been addressed in previous studies~\cite{9412307, mahbod2021pollen, MAHBOD2020105475}. However, in this study, we investigated its effect on the retrieval performance of both CNNs and foundation models. The reported ACC@1 results in Figures~\ref{fig:BreastMNIST} to \ref{fig:PathMNIST} for 2D datasets indicate an overall slight increase in the ACC@1 performance with increasing image size, but even for smaller sizes, the performance remains competitive. These figures also show that the models delivered the best overall ACC@1 performance for the PathMNIST dataset, while the worst performance was observed for the RetinaMNIST dataset. All foundation models (except MedCLIP) achieved at least a 68.9\% ACC@1 score for all sub-datasets (except RetinaMNIST) across all image sizes. The poor performance on the RetinaMNIST dataset in our experiments could be related to the low number of training and test images, as RetinaMNIST is the second smallest dataset among the sub-datasets we selected from the MedMNIST V2.

\begin{table*}[]
	\caption[]{Average results across all 2D datasets and all sizes (\%). For each evaluation metric, the best results are shown in \textbf{bold} and the second-best results are shown with \underline{underline}.
	}
	\label{datasets}
	\begin{tabular}{lccccc}
		\hline
		\textbf{Model} &
		\textbf{mAP@5} & \textbf{mMV@5} & \textbf{ACC@1} & \textbf{ACC@3}  & \textbf{ACC@5} \\
		\hline
		\textbf{VGG19}                &77.69 		  		&77.13	        	&72.27      		&86.70				&87.39          \\
		\textbf{ResNet50}             &79.00 		  		&78.67	        	&73.58         		&87.99				&92.57           \\
		\textbf{DenseNet121}          &79.10 		  		&78.53        		&74.07	        	&88.04				&92.66            \\
		\textbf{EfficientNetV2M}      &74.95 	      		&74.07	        	&68.78	        	&85.05				&90.31           \\	\hline
		\textbf{MedCLIP}              &76.22 		  		&76.10	        	&70.01	        	&86.02				&90.99             \\
		\textbf{BiomedCLIP}           &\underline{83.18} 	&\underline{81.87}	&\underline{79.16}  &\underline{90.44}	&\textbf{94.08}    \\
		\textbf{OpenCLIP}             &82.65 		  		&81.46	        	&78.44	        	&\textbf{90.46}		&\underline{94.04} \\
		\textbf{CONCH}                &82.65				&81.24				&78.46				&89.65				&93.45\\
		\textbf{UNI}                  &\textbf{83.50} 		&\textbf{82.18}		&\textbf{79.45}		&90.33				&93.94            \\

		\hline
	\end{tabular}
	\label{tab:2d}
\end{table*} 

\begin{figure}[]
	\centering
	\begin{tabular}{c}
		
		\includegraphics[width=\linewidth]{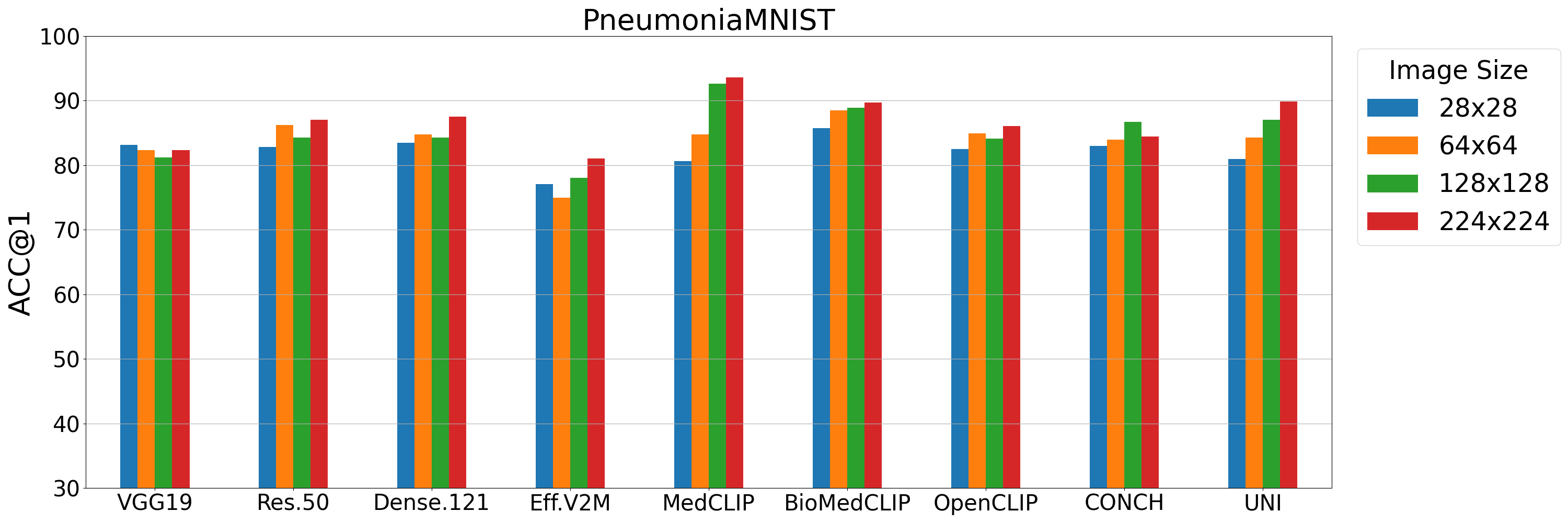} \\
	\end{tabular}
	\caption{ACC@1 performance based on different image sizes for the PneumoniaMNIST dataset. Res.50 = ResNet50, Dense.121 = DenseNet 121, Eff.V2M = EfficientNetV2M.}
	\label{fig:PneumoniaMNIST}
\end{figure}

\subsection{3D Results}
\label{sec:3D Results}
Although the performance difference between foundation models and CNNs is more evident for 2D datasets, for 3D datasets, the performances of both are competitive. The results in Table~\ref{tab:3d} show that CONCH delivers the best overall retrieval performance across all models, but only 2\% better compared to the best CNN model (DenseNet121) based on the ACC@1 score. For other evaluation indices, the differences are even smaller (1.15\%, -0.68\%, 1.89\%, 0.49\% for mAP@5, mMV@5, ACC@3, and ACC@5, respectively). However, it is still interesting that CONCH, which is only trained on histological images, is able to perform well on 3D CT and electron microscopy images. As can be seen in Tables S1 and S2 and Figure~\ref{fig:AdrenalMNIST3D} and Figure~\ref{fig:SynapseMNIST3D}, CONCH performs better overall on the SynapseMNIST3D dataset, as electron microscopy images are more similar to histological images compared to CT images. 

Investigating the effect of image size on ACC@1 performance in Figure~\ref{fig:AdrenalMNIST3D} and Figure~\ref{fig:SynapseMNIST3D} is challenging, as only two sizes are available for 3D datasets. However, the results in these figures show that for the best CNN model for 3D datasets (DenseNet121), the increase in size has led to an ACC@1 increase for both 3D datasets, while for the best foundation model for 3D datasets (CONCH), increasing the size has no effect on the AdrenalMNIST3D dataset but improves ACC@1 performance for the SynapseMNIST3D dataset.

\subsection{Merged Results}
\label{sec:Merged Results}
\begin{figure}[]
	\centering
	\begin{tabular}{c}
		
		\includegraphics[width=\linewidth]{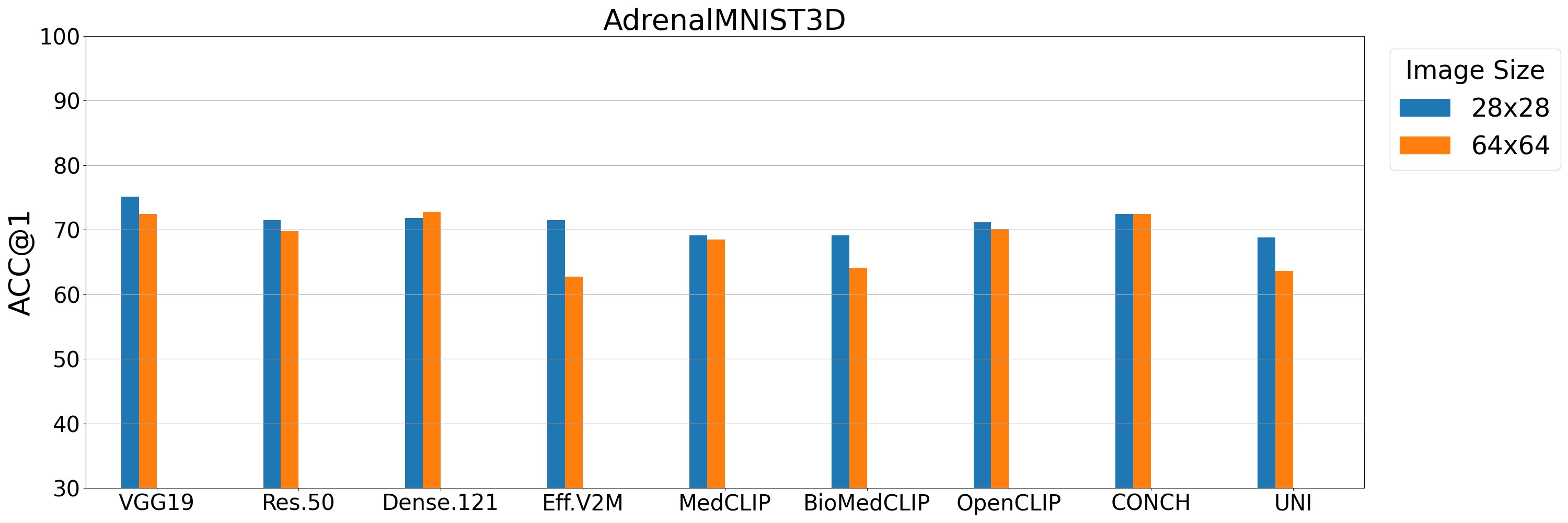} \\
	\end{tabular}
	\caption{ACC@1 performance based on different image sizes for the AdrenalMNIST3D dataset. Res.50 = ResNet50, Dense.121 = DenseNet 121, Eff.V2M = EfficientNetV2M.}
	\label{fig:AdrenalMNIST3D}
\end{figure}

\begin{figure}[h!]
	\centering
	\begin{tabular}{c}
		
		\includegraphics[width=\linewidth]{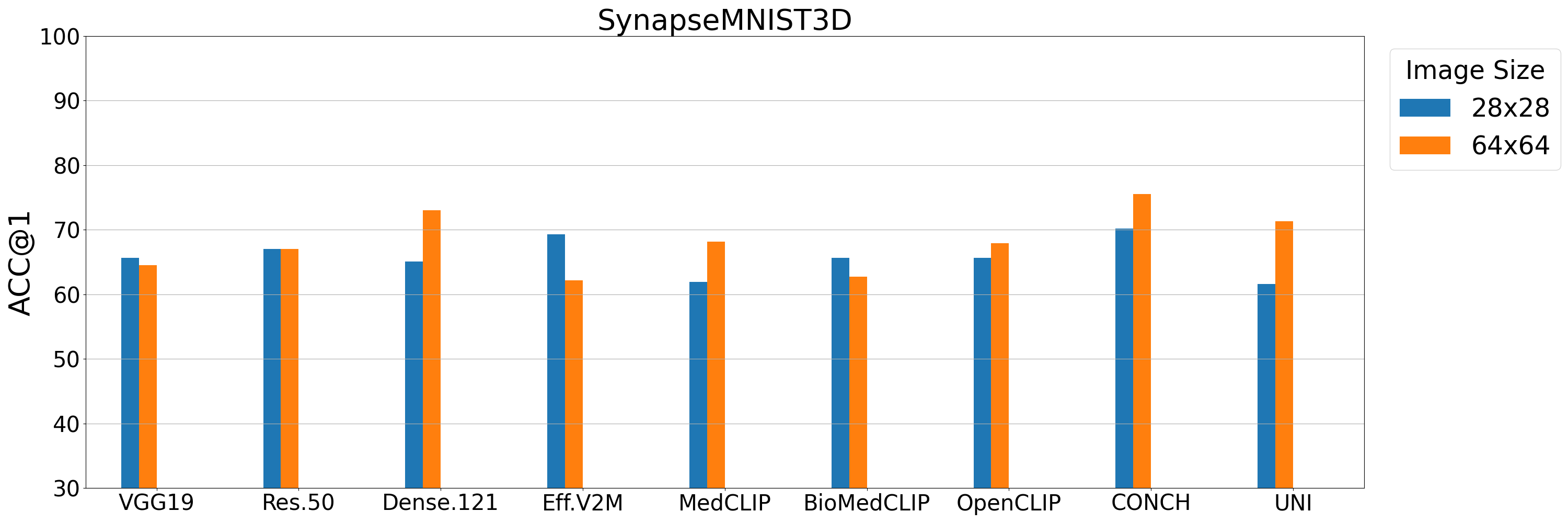} \\
	\end{tabular}
	\caption{ACC@1 performance based on different image sizes for the SynapseMNIST3D dataset. Res.50 = ResNet50, Dense.121 = DenseNet 121, Eff.V2M = EfficientNetV2M.}
	\label{fig:SynapseMNIST3D}
\end{figure}

Considering all ACC@1 results from Figure~\ref{fig:BreastMNIST} to Figure~\ref{fig:SynapseMNIST3D}, it can be observed that among all the 2D and 3D datasets, the models deliver the most consistent performance on the PneumoniaMNIST dataset, with almost all values above 80\%. In contrast, the largest variation in performance was observed in the RetinaMNIST dataset, where values ranged from 33.00\% to 46.77\%.

\begin{table*}[]
	\caption[]{Average results across all 3D datasets and all sizes (\%). For each evaluation metric, the best results are shown in \textbf{bold} and the second-best results are shown with \underline{underline}.
	}
	\label{datasets}
	\begin{tabular}{lccccc}
		\hline
		\textbf{Model} &
		\textbf{mAP@5} & \textbf{mMV@5} & \textbf{ACC@1} & \textbf{ACC@3}  & \textbf{ACC@5} \\
		\hline
		\textbf{VGG19}                &77.49				&73.29				&69.44				&\underline{91.49}	&\underline{96.02}           \\
		\textbf{ResNet50}             &77.42				&74.64   			&68.85				&90.83				&94.97  \\
		\textbf{DenseNet121}          &\underline{77.71}	&\textbf{75.70}		&\underline{70.68}	&89.20				&94.00     \\
		\textbf{EfficientNetV2M}      &74.51				&71.29				&66.44				&88.29				&94.36  \\	\hline
		\textbf{MedCLIP}              &75.95				&73.22				&66.93				&89.40				&95.28  \\
		\textbf{BiomedCLIP}           &75.68				&73.58				&65.41				&89.61				&94.53  \\
		\textbf{OpenCLIP}             &77.08				&73.72				&68.70				&\textbf{92.11}		&\textbf{96.55}  \\
		\textbf{CONCH}                &\textbf{78.86}		&\underline{75.02}	&\textbf{72.68}		&91.09				&94.49\\
		\textbf{UNI}                  &75.94				&72.95				&66.36				&90.29				&95.66  \\

		\hline
	\end{tabular}
	\label{tab:3d}
\end{table*}




\begin{figure}[]
	\centering
	\begin{tabular}{ccccc}
		
		\includegraphics[width=3cm]{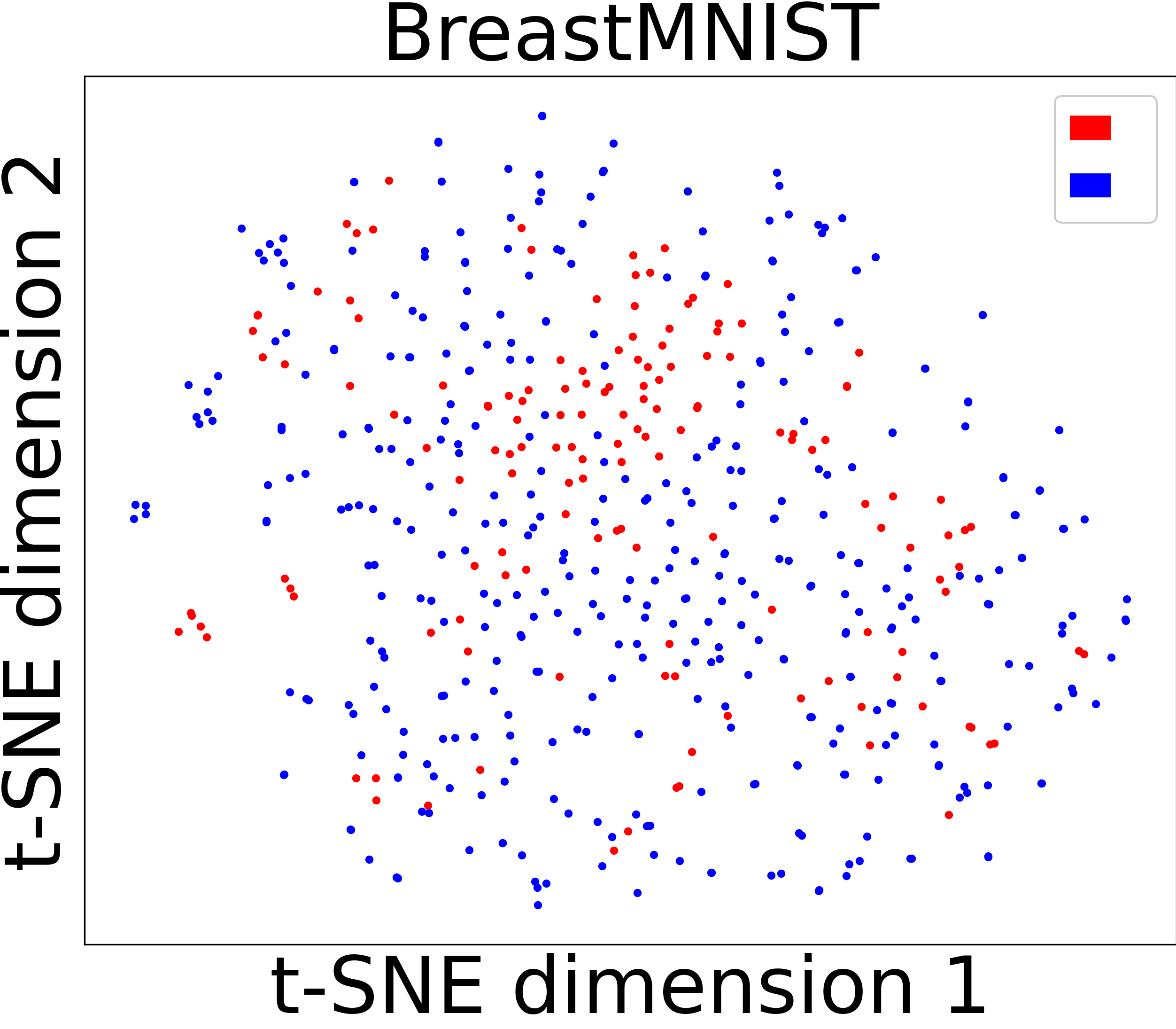} &
		\includegraphics[width=3cm]{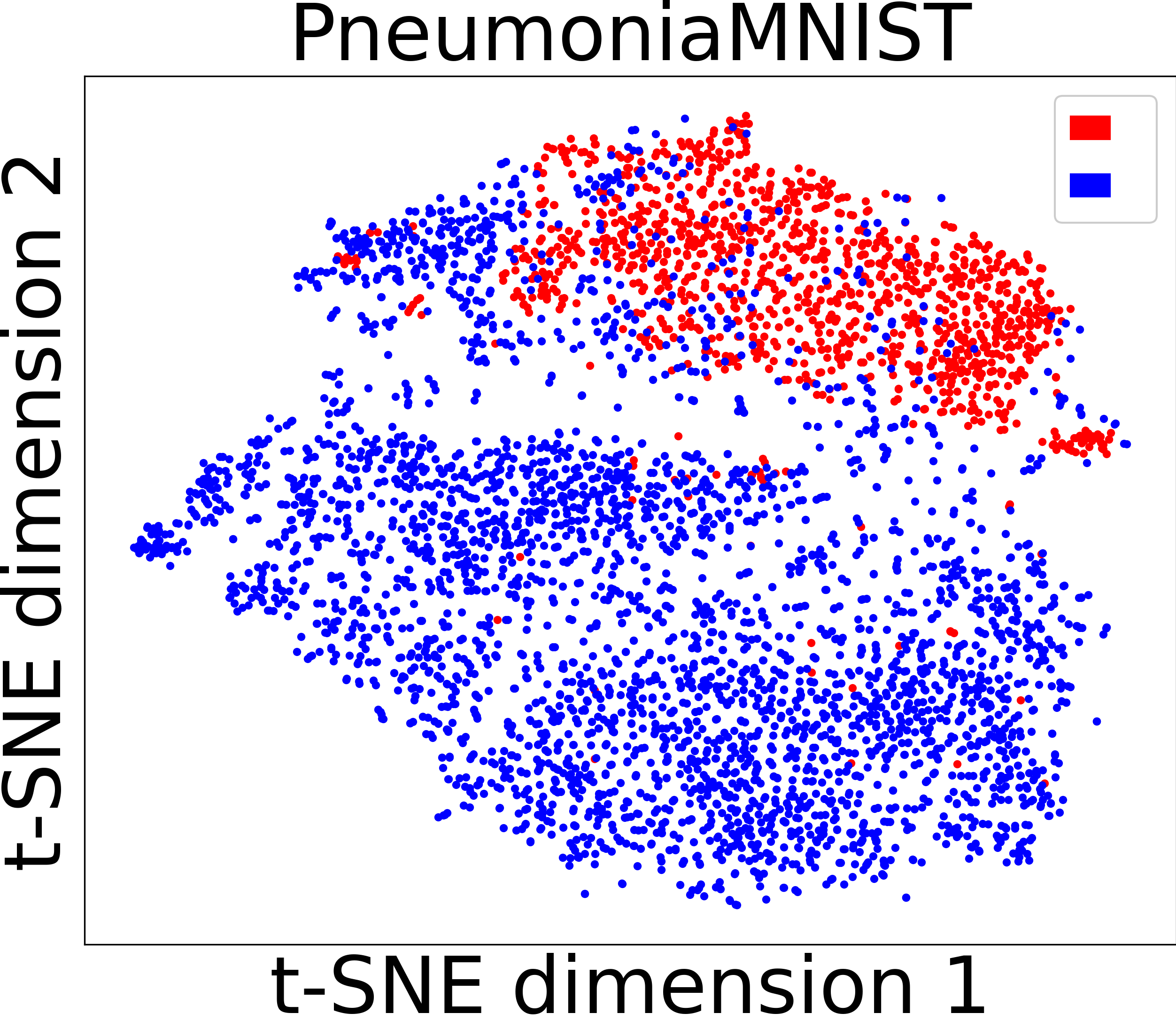} &
		\includegraphics[width=3cm]{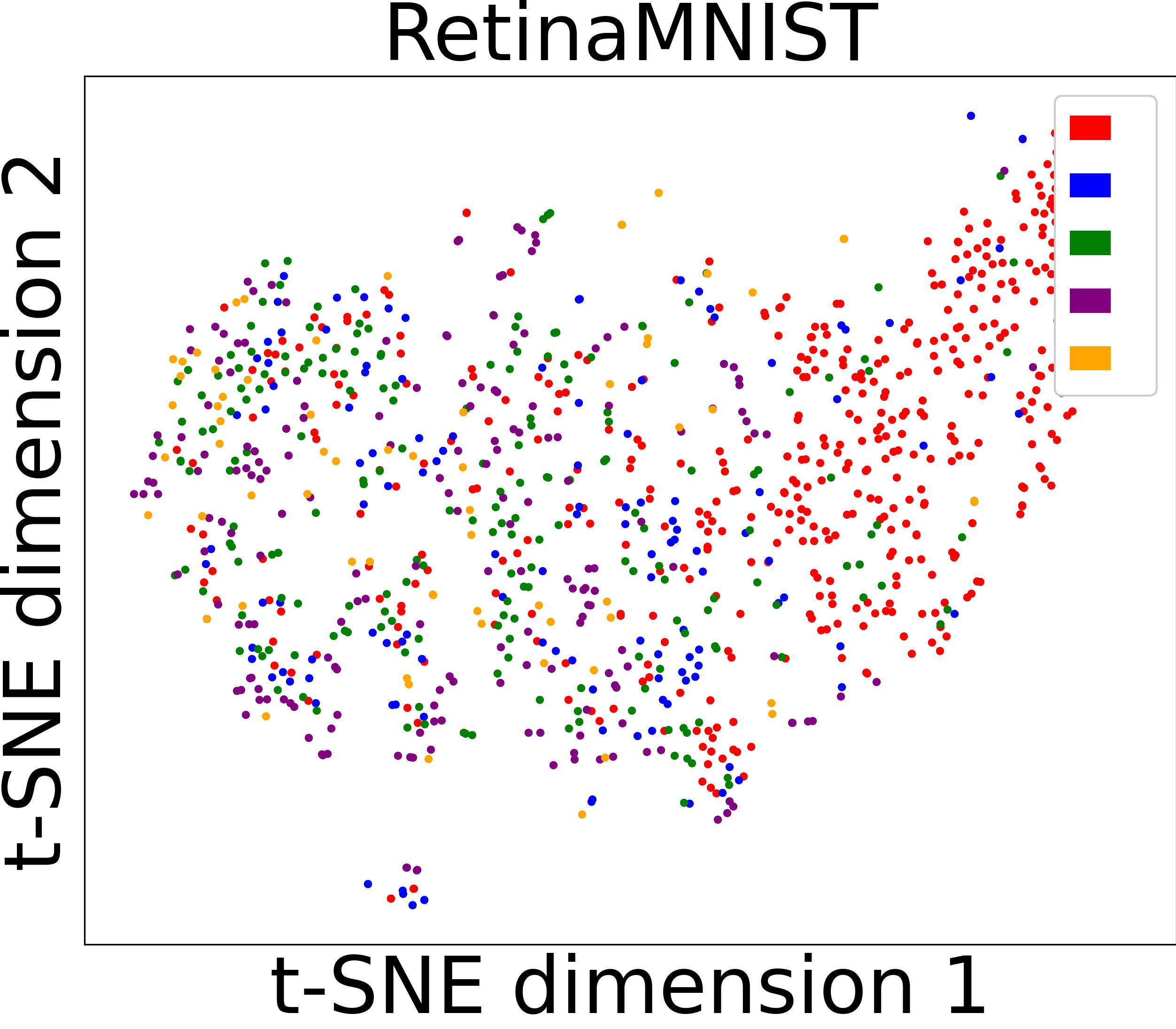} &
		\includegraphics[width=3cm]{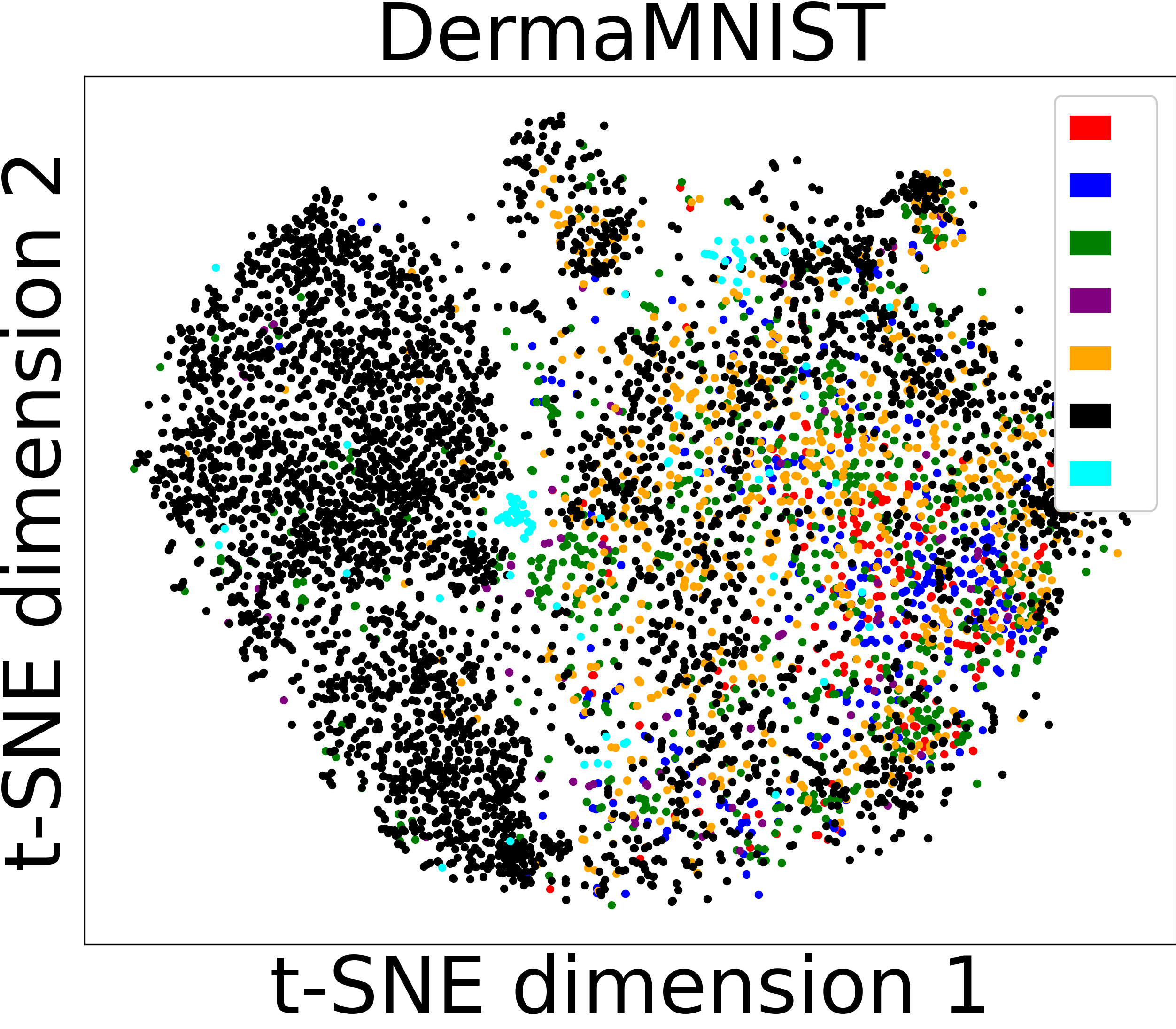}  \\
	
		\\
		\includegraphics[width=3cm]{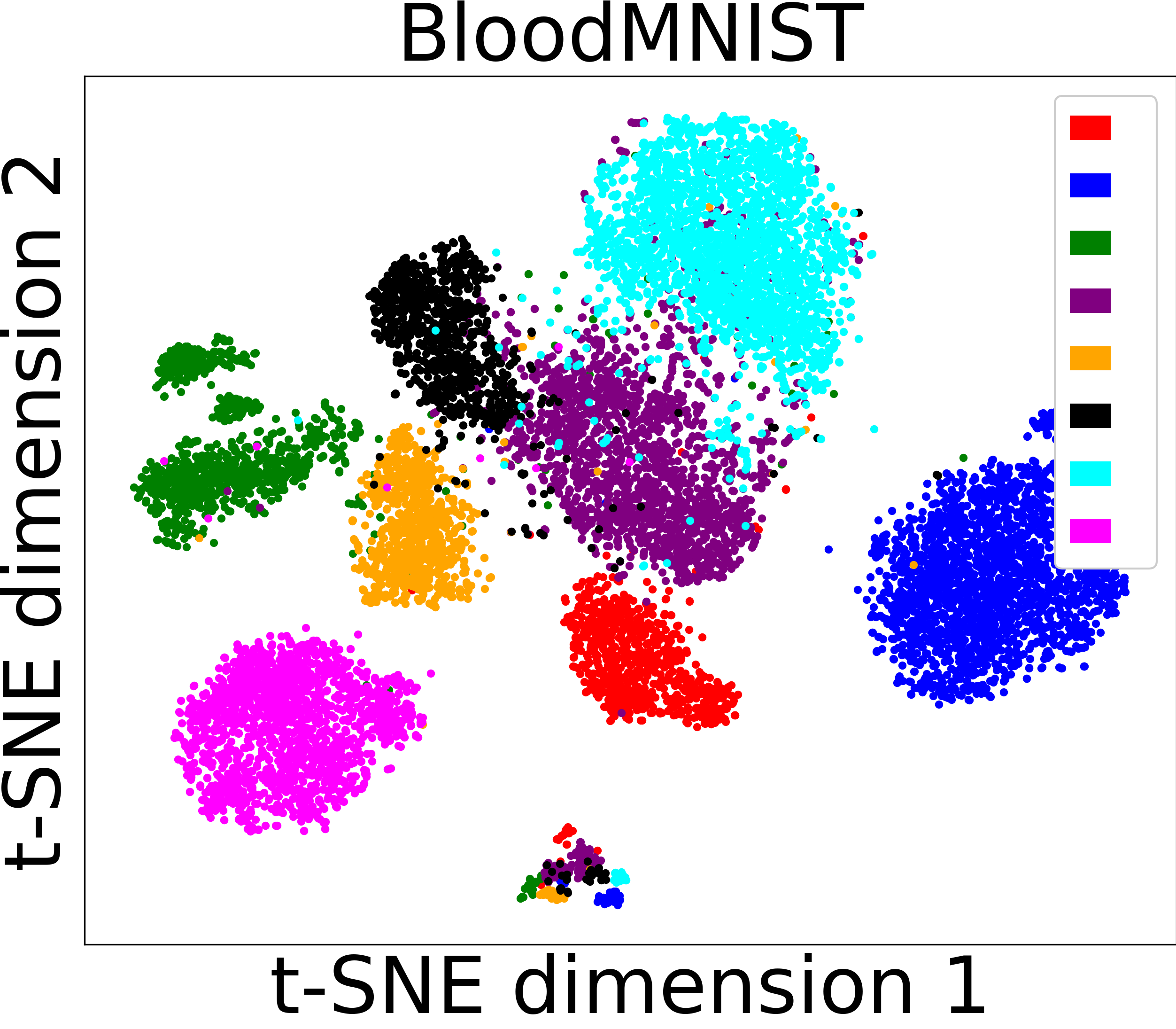} &
		\includegraphics[width=3cm]{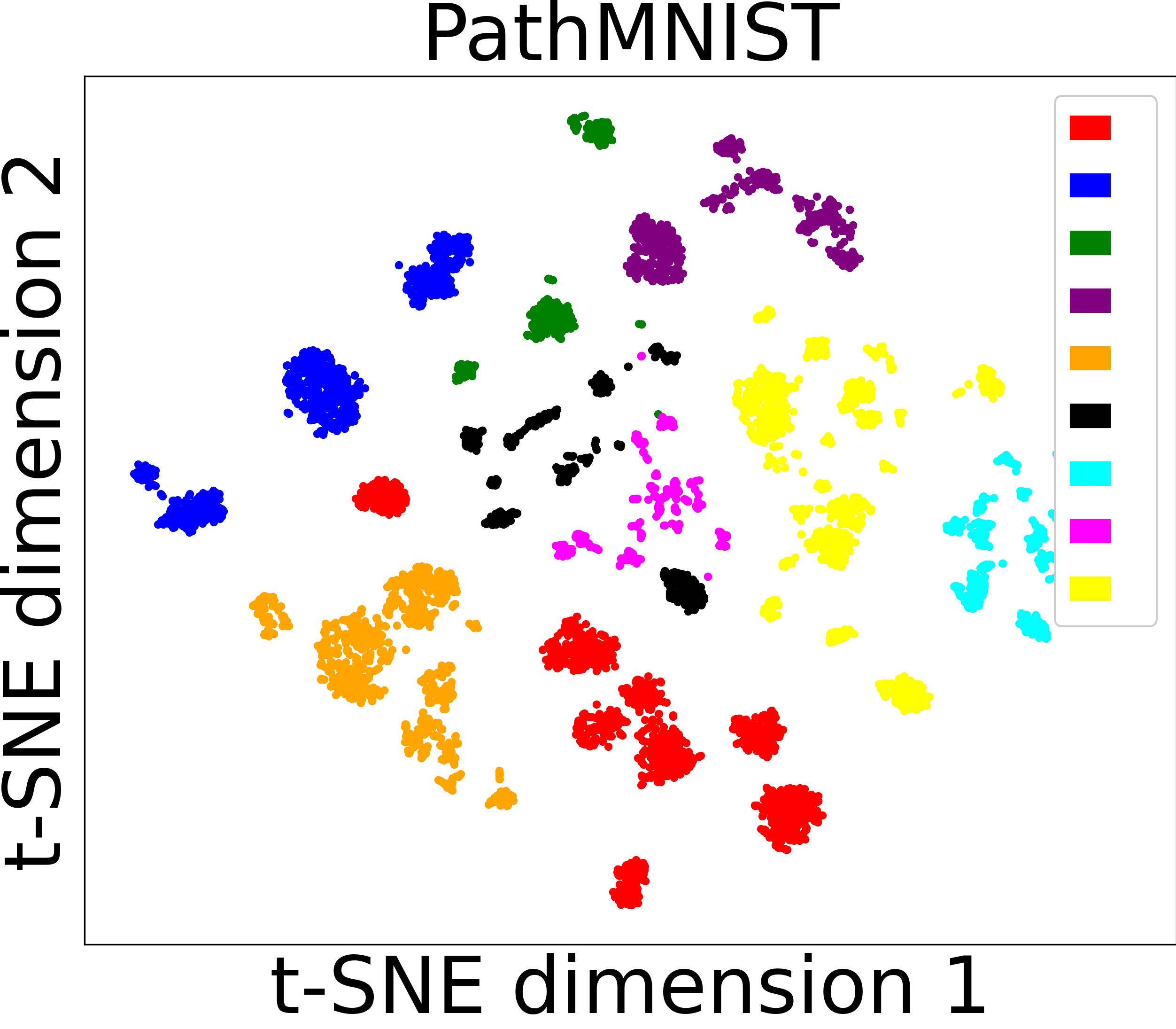} &
		\includegraphics[width=3cm]{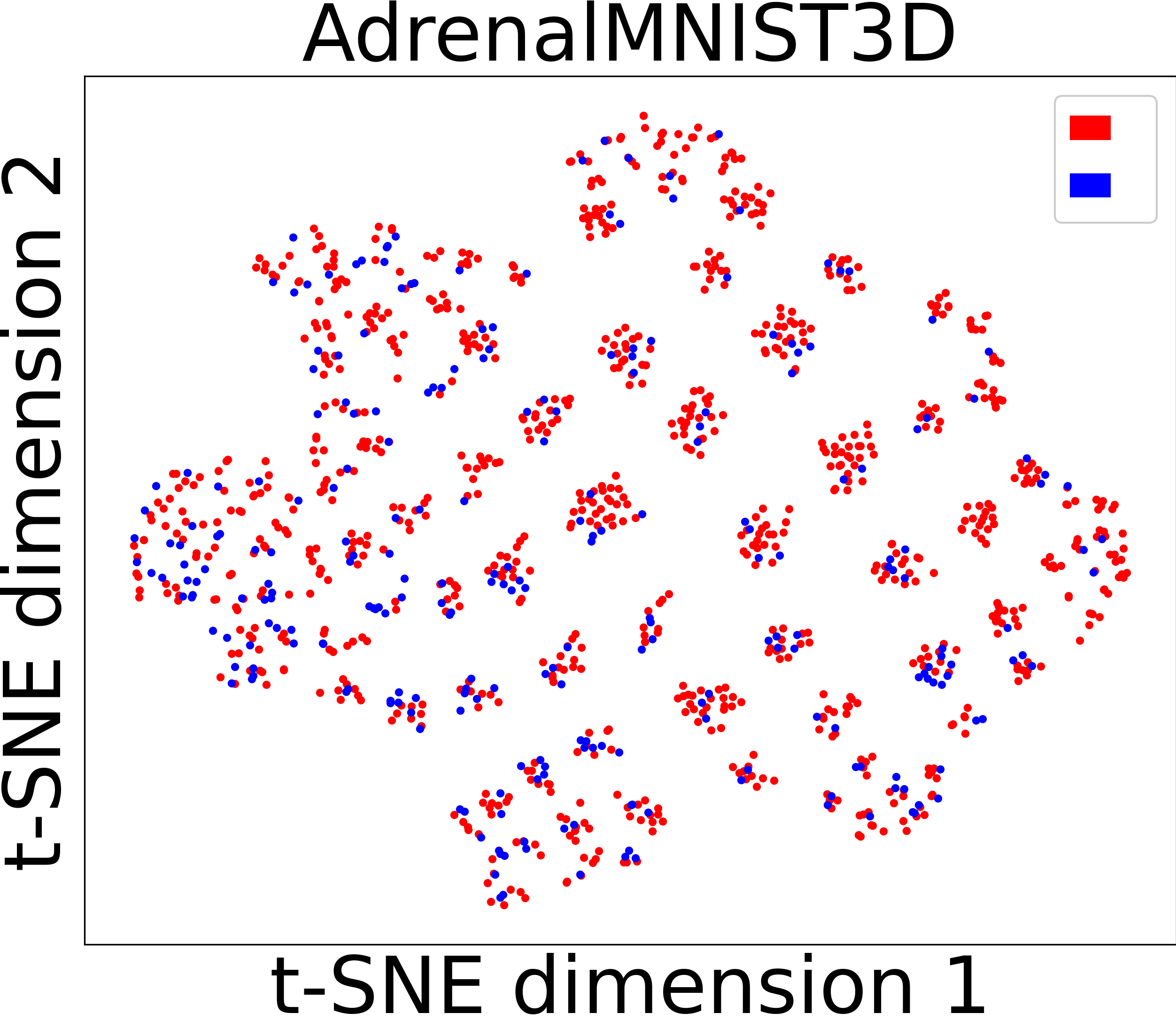} &
		\includegraphics[width=3cm]{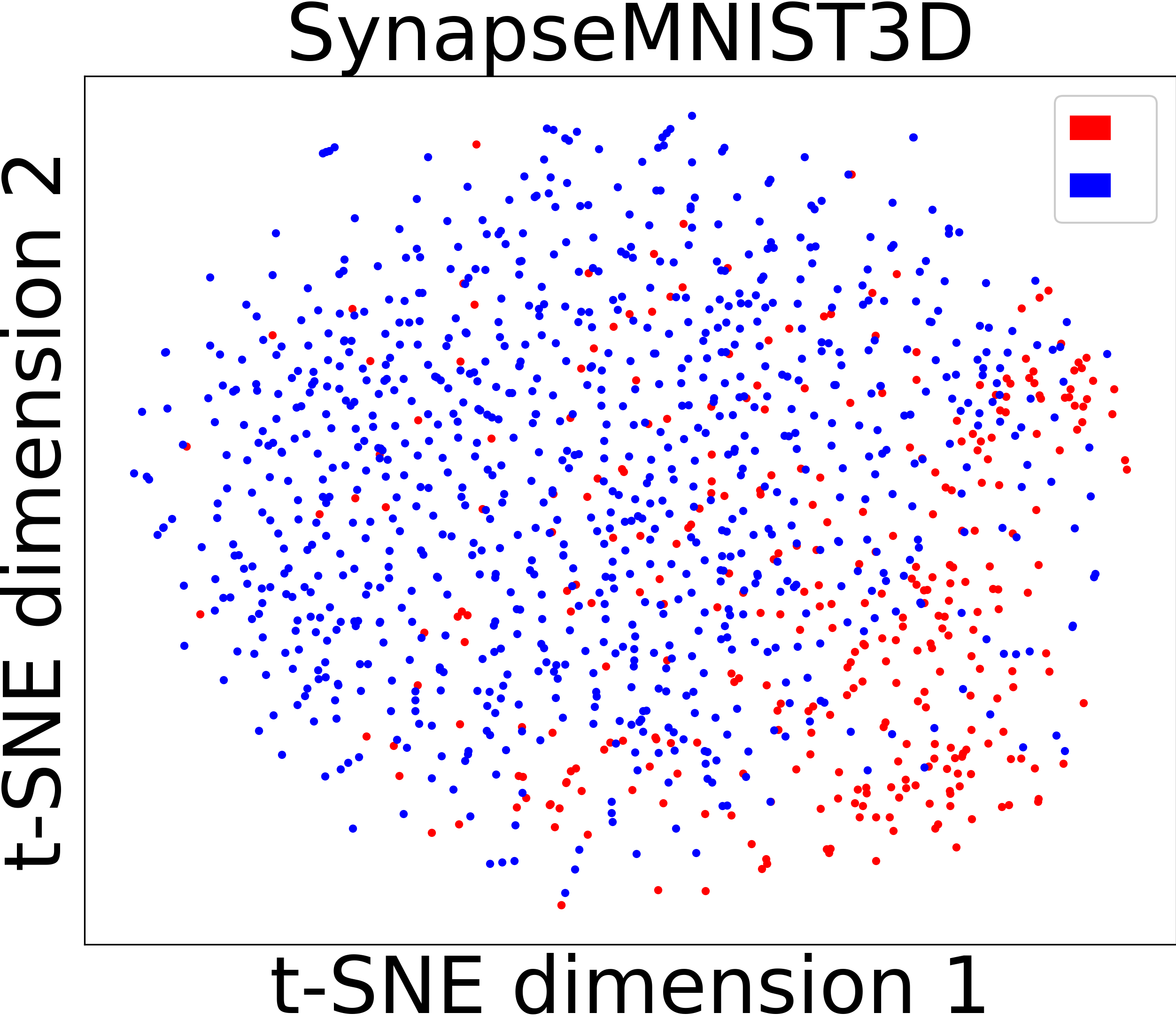}

	\end{tabular}
	\caption{t-SNE~\cite{JMLR:v9:vandermaaten08a} plots of the extracted features by the UNI model~\cite{Chen2024}. For all datasets except PathMNIST, the features from the training set were used to better visualize the clustering. For all 2D datasets, 224$\times$224 images were used, while for 3D datasets, 64$\times$64 images were employed. The color legend on the top right of each subfigure represents the classes within the corresponding datasets.}
	\label{fig:tsne}
\end{figure}

To demonstrate the capabilities of the extracted features from one of the best-performing models (UNI), we show the two-dimensional feature maps using the t-SNE approach for all eight sub-datasets in Figure~\ref{fig:tsne}. We use the training set for plotting to better illustrate the clustering, except for the PathMNIST dataset, where test set images were used due to the high number of images in the training set, which made the clusters too dense for visualization. As evident from the figure, features of different classes are well distinguishable for some datasets (e.g., BloodMNIST and PathMNIST), where the retrieval performance is also high. However, for other datasets (e.g., RetinaMNIST), where the retrieval performance is lower, the clusters are not as well separated.

We assessed model sensitivity and robustness to image size variations by computing the range of ACC@1 values (i.e., the difference between the highest and lowest values) for each dataset. To determine overall trends, we averaged these ranges across all eight datasets. The results are reported in Table~\ref{tab:range}.
We define a model whose performance remains relatively stable (i.e., a small range) as “robust” with respect to image-size variation, while a model whose performance fluctuates a lot (i.e., a large range) is considered “sensitive” to image size.
As evidenced by the results, OpenCLIP is the most overall robust model, while MedCLIP, which was the worst-performing foundation model in our experiments, is also the most overall sensitive model to changes in image size.

\begin{table*}[]
	\caption[]{The most robust and most sensitive models to changes in image size based on each dataset and the overall average. To determine the range, ACC@1 scores were used.  In the dataset names in the first column, "MNIST" is removed (e.g., Breast refers to BreastMNIST, and Adrenal3D refers to AdrenalMNIST3D). Res.50 = ResNet50, Eff.V2M = EfficientNetV2M.
	}
	\label{datasets}
	\begin{tabular}{lccccc}
		\hline
		\textbf{Dataset} &
		  \textbf{Most Robust} & \textbf{Range (\%)} & \textbf{Most Sens.}  & \textbf{Range (\%)} \\
		\hline
		\textbf{Breast}                & OpenCLIP	 &	2.56    & CONCH	    & 11.54	    \\
		\textbf{Pneumonia}             & VGG19	     & 1.92	    & MedCLIP	& 12.98	    \\
		\textbf{Retina}                & CONCH	     &	2.50	& MedCLIP	& 10.50	    \\
		\textbf{Derma}                 & CONCH       & 5.34	    & UNI		& 11.22		\\	
		\textbf{Blood}                 & CONCH       & 8.80	    & MedCLIP	& 30.46		\\
		\textbf{Path}                  & UNI         &	5.50	& Res.50	& 15.44	    \\
		\textbf{Adrenal3D}             & CONCH       &	0.00	& Eff.V2M	& 8.72	    \\
		\textbf{Synapse3D}             & Res.50    & 0.00	    & UNI 	    & 9.66		\\ \hline
		\textbf{All}                   & OpenCLIP    & 5.76	    & MedCLIP	& 10.36		\\

		\hline
	\end{tabular}
	\label{tab:range}
\end{table*}

To investigate the speed of the models, we report the dataset creation time (referred to as training time) and testing time for the largest common image size in the 2D and 3D datasets (i.e., 64) in Table~\ref{tab:time_CNN} and Table~\ref{tab:time_foundation} for CNN and foundation models, respectively. 

As the results indicate, for the 2D datasets, DenseNet121 and ResNet50 are the fastest CNN models, while MedCLIP and UNI are the fastest foundation models. Regarding ACC@1, MedCLIP delivered the weakest performance across all foundation models. However, UNI, which is the second fastest foundation model, is almost as fast as the DenseNet121 model. If we consider only the testing time (which is more important in real-world applications), UNI is nearly as fast as DenseNet121 and significantly faster in the testing phase for the PathMNIST dataset (17.82 vs. 40.48 minutes). 

For the 3D datasets, VGG19 and ResNet50 are the fastest CNN models, while BioMedCLIP and MedCLIP are the fastest foundation models. Comparing the fastest CNN and foundation models for 3D datasets, it is evident that BioMedCLIP is significantly faster than VGG19 by a large margin in both training and testing time.

Considering the data sizes, it is clear that performing retrieval on 3D datasets is much more time-intensive compared to 2D datasets. For example, for the UNI model, the normalized training and testing times (derived by dividing training/testing times in Table~\ref{tab:datasets} by the total number of training/testing images in Table~\ref{tab:time_foundation}) for the PathMNIST dataset (the largest 2D dataset) and the Synapse3D dataset (the largest 3D dataset) are 0.00021/0.00248 and 0.02209/0.02307, respectively.

We acknowledge that image size affects the CNN feature extraction speed. However, the test phase time consists of two steps: feature extraction from test images and calculating the cosine similarity index between the test image features and the stored database features. While image size influences feature extraction time, it does not impact the speed of calculating the cosine similarity index. Our experiments show that between 90\% and 99.5\% of the test time is spent on computing the cosine similarity index, while only 0.5\% to 10\% is related to feature extraction from test images. For foundation models, since images are resized to the compatible input size of the models before processing (i.e., $224\times224$ pixels), image size does not affect the test time speed. Therefore, image size has only a very slight influence on the overall test time speed.

\begin{table*}[]
	\caption[]{Database creation (training) time and testing time for the convolutional neural network models based on each dataset. 64$\times$64 and 64$\times$64$\times$64 images were used to report the results. The best and the second best results (based on shortest average training and testing times) for each dataset are highlighted in \textbf{bold} and with \underline{underline}underline, respectively. All results are reported in minutes. Res.50 = ResNet50, Dense121 = DenseNet121, Eff.V2M = EfficientNetV2M.
	}
	\label{datasets}
	\begin{tabular}{lccccc}
		\hline
		\textbf{Dataset} &
		 \textbf{\begin{tabular}[c]{@{}c@{}}VGG19 \\ \small train/test\end{tabular}} & \textbf{\begin{tabular}[c]{@{}c@{}}Res.50 \\ \small train/test \end{tabular}} & \textbf{\begin{tabular}[c]{@{}lc@{}}Dense.121 \\ \small train/test \end{tabular}}  & \textbf{\begin{tabular}[c]{@{}c@{}}Eff.V2M \\  \small train/test \end{tabular}} \\
		\hline
		\textbf{Breast}     &0.06/0.02             &\underline{\textbf{0.03/0.01}}	 &\underline{\textbf{0.03/0.01}}  &0.07/0.01 \\
		\textbf{Pneumonia}  &0.51/0.22             &\underline{0.19/0.22}            &\textbf{0.18/0.18}              &0.32/0.20  \\
		\textbf{Retina}     &0.12/0.07 	           &\underline{\textbf{0.05/0.04}} 	 &\underline{\textbf{0.05/0.04}}  &0.11/0.05  \\
		\textbf{Derma}      &0.78/0.97             &\underline{0.27/0.95} 		     &\textbf{0.26/0.84}              &0.48/0.92  \\	
		\textbf{Blood}      &1.33/2.59             &\underline{0.46/2.74} 		     &\textbf{0.42/2.52}              &0.79/2.73 \\
		\textbf{Path}       &10.29/37.41           &\underline{3.45/44.14} 		     &\textbf{3.22/40.48} 		      &5.45/42.01 \\
		\textbf{Adrenal3D}  &\textbf{41.37/10.49}  &\underline{46.52/11.92}	         &52.46/13.31	                  &68.65/17.40    \\
		\textbf{Synapse3D}  &\textbf{42.81/12.33}  &\underline{47.94/14.30}	         &54.97/15.99	                  &72.06/20.84	  \\
		
		\hline
	\end{tabular}
	\label{tab:time_CNN}
\end{table*}

\begin{table*}[]
	\caption[]{Database creation (training) time and testing time for the foundation models based on each dataset. 64$\times$64 and 64$\times$64$\times$64 images were used to report the results. The best and the second best results (based on shortest average training and testing times) for each dataset are highlighted in \textbf{bold} and with \underline{underline}underline, respectively. All results are reported in minutes. Med = MedCLIp, BioMed = BioMedCLIP, Open = OpenCLIP.
	}
	\label{datasets}
	\begin{tabular}{lccccc}
		\hline
		\textbf{Dataset} &
		\textbf{\begin{tabular}[c]{@{}c@{}}Med \\ \small train/test \end{tabular}} & \textbf{\begin{tabular}[c]{@{}c@{}}BioMed \\ \small train/test\end{tabular}} & \textbf{\begin{tabular}[c]{@{}c@{}}Open \\ \small train/test\end{tabular}}  & \textbf{\begin{tabular}[c]{@{}c@{}}CONCH \\ \small train/test\end{tabular}} & \textbf{\begin{tabular}[c]{@{}c@{}}UNI \\ \small train/test \end{tabular}}\\
		\hline
		\textbf{Breast}     &\textbf{0.03/0.01}      &0.15/0.04	        	&0.12/0.05  	&0.17/0.05	       &\underline{0.12/0.03} \\
		\textbf{Pneumonia}  &\textbf{0.18/0.05}      &1.14/0.30	         	&0.95/0.82 		&1.45/0.23 		   &\underline{0.95/0.20} \\
		\textbf{Retina}     &\textbf{0.05/0.02}      &0.27/0.12	            &0.23/0.18 	    &0.33/0.12 	       &\underline{0.22/0.09}  \\
		\textbf{Derma}      &\textbf{0.25/0.22}      &1.72/1.24		        &1.72/1.24		&2.19/0.80 	       &\underline{1.46/0.75} \\	
		\textbf{Blood}      &\textbf{0.43/0.56}	     &3.03/2.95		        &2.46/10.32 	&3.77/1.52 	       &\underline{2.52/1.78} \\
		\textbf{Path}       &\textbf{3.01/8.53}	     &23.23/35.45        	&23.23/35.45	&28.19/10.02 	   &\underline{18.59/ 17.82} \\
		\textbf{Adrenal3D}  &\underline{21.76/5.66}	 &\textbf{21.27/ 5.42}	&45.21/ 15.98	&29.92/7.57	       &26.35/6.84    \\
		\textbf{Synapse3D}  &\underline{23.63/6.88}  &\textbf{21.83/6.39}	&46.56/ 18.65	&30.72/8.90	       &27.17/8.12 \\
		
		\hline
	\end{tabular}
	\label{tab:time_foundation}
\end{table*}

\subsection{Overall Discussion on the Advantages and Disadvantages of the CNNs and Foundation Models for CBMIR }
\label{sec:overall}
Below, we outline the overall strengths and weaknesses of the CNNs and foundation models used in this study based on our findings.

We evaluated four commonly used CNN architectures: VGG19, ResNet50, DenseNet121, and EfficientNetV2M. Their overall advantages can be listed as follows: 
\begin{itemize}
	\item Faster database creation for 2D datasets 
	\item Comparable retrieval performance for 3D images
	\item Lower computational overhead for training (CNN models are pre-trained on the ImageNet dataset, which is typically smaller than the datasets used to train foundation models)
\end{itemize}
and their disadvantages can be listed as follows:
\begin{itemize}
    \item Inferior performance of retrieval task for 2D datasets
    \item Slower database creation and testing for 3D datasets

\end{itemize}

Regarding the foundation models we evaluated five models: MedCLIP, BioMedCLIP, OpenCLIP, CONCH, and UNI. Their overall advantages can be listed as follows: 
\begin{itemize}
    \item Superior performance in 2D medical image retrieval
    \item Stronger feature representations and better generalization across medical modalities
     \item Faster database creation and testing for 3D datasets, and nearly as fast as CNNs for testing in 2D datasets
     \item Better performance on in-domain datasets (e.g., UNI for PathMNIST)
     \item Robust performance across different image sizes (with the exception of the MedCLIP model)
\end{itemize}
and their disadvantages can be listed as follows:
\begin{itemize}
    \item Sub-optimal performance for 3D image retrieval, comparable to CNN performance
    \item Slower database creation for 2D datasets 
     \item Higher computational cost in training (foundation models are typically trained on millions or billions of data samples)
	
\end{itemize}

\subsection{Limitations and Future Works}
There are certain limitations with the current study. First, although we utilized several well-known CNNs and foundation models, this study can be further expanded by incorporating additional models. It should be noted that in this research, we focused on foundation models that have demonstrated excellent performance in downstream computer vision tasks and excluded others that have already shown inferior performance in previous studies on medical image retrieval. For CNN models, we employed well-established pre-trained models based on the ImageNet dataset and used them as benchmarks. However, certain CNN-based models trained on medical images such as ~\cite{XIE2024103023, 10242007, 10605840, 10617802} can also be used, and their performance can be compared to foundation models. However, this approach has become less common in recent years due to the superior performance of ViT-based models over CNNs for pre-training.

Another aspect that can be further investigated in future studies is exploring a more computationally efficient fusion scheme to merge features from 2D slices for 3D datasets. Although some efforts, such as slice sampling or slice-based retrieval and voting, have been explored in previous studies~\cite{10635170, jush2024content} and we also tried well-known feature reduction methods (Table S5 and Table 6 in the supplementary materials), there is still room for improvement in this area. 

2D slicing has another inherent issue with the loss of spatial information between slices. However, to utilize the CNNs and foundation models used in this study, which were initially trained on 2D images, we extracted compatible 2D images from the 3D volume to serve as model inputs. Although the loss of spatial information between slices is inevitable, it should be noted that this does not necessarily lead to inferior performance. For example, as demonstrated in other computer vision tasks such as segmentation, 2D/2.5D U-Net models that do not preserve spatial information between slices have achieved comparable segmentation results to 3D U-Net models that retain spatial information between slices for certain medical imaging datasets~\cite{isensee2021nnu}. 

Although in this study we investigated the effect of image size on the retrieval performance, the maximum image size we examined was 224$\times$224 pixels, which is the largest size provided in the latest version of the MedMNIST V2 dataset. Larger image sizes can also be explored in future studies. However, it is important to note, as we have shown in this work, that even smaller image sizes can still deliver competitive retrieval performance. Additionally, as demonstrated in the BioMedCLIP study~\cite{zhang2023large}, using larger image sizes does not necessarily result in superior performance.

Finally, similar comprehensive studies can be conducted for other downstream medical computer vision tasks to investigate the performance and generalization capabilities of foundation models in other domains than medical image retrieval as performance may vary for different tasks~\cite{10205297}.

\section{Conclusion}
\label{sec:Conclusion}
In this study, a comprehensive analysis of CBMIR was conducted using pre-trained CNNs and foundation models. The focus was on eight types of 2D and 3D medical images from the MedMNIST V2 dataset, with the aim of evaluating retrieval performance across different models and image sizes. Our results show that foundation models (especially UNI and CONCH) generally deliver excellent performance for all datasets, while the superiority was shown to be more evident for 2D datasets. This study also explored the effect of image size on retrieval performance and revealed that larger image sizes generally led to slight improvements in retrieval performance. However, the performance of smaller images remained competitive. Finally, we pointed out the limitations of the current study and opened the door for future research into improving feature extraction techniques, optimizing retrieval for 3D medical images, and expanding the use of foundation models in other medical imaging tasks.

\section*{Acknowledgments}
We would like to thank the NVIDIA corporation for their generous GPU donation. This research did not receive any specific grant from funding agencies in the public, commercial, or not-for-profit sectors.
	
\section*{Conflict of Interest}
The authors declare that they have no known competing financial interests or personal relationships that could have appeared to influence the work reported in this paper.
	
\section*{Credit Authorship Contribution Statement}
Amirreza Mahbod: Conceptualization, Methodology, Investigation, Validation, Software, Formal analysis, Writing original draft. 
Nematollah Saeidi: Methodology, Investigation, Software. 
Sepideh Hatamikia: Investigation, Validation.  
Ramona Woitek: Investigation, Validation, Supervision. 
All authors read and agreed to the current version of the manuscript.

\bibliographystyle{elsarticle-num}
\bibliography{MIR_PTM.bib}

\begin{thebibliography}{10}
\expandafter\ifx\csname url\endcsname\relax
  \def\url#1{\texttt{#1}}\fi
\expandafter\ifx\csname urlprefix\endcsname\relax\def\urlprefix{URL }\fi
\expandafter\ifx\csname href\endcsname\relax
  \def\href#1#2{#2} \def\path#1{#1}\fi

\bibitem{Vishraj2022}
R.~Vishraj, S.~Gupta, S.~Singh, A comprehensive review of content-based image
  retrieval systems using deep learning and hand-crafted features in medical
  imaging: Research challenges and future directions, Computers and Electrical
  Engineering 104 (2022).
\newblock \href
  {https://doi.org/https://doi.org/10.1016/j.compeleceng.2022.108450}
  {\path{doi:https://doi.org/10.1016/j.compeleceng.2022.108450}}.

\bibitem{5580313}
G.~Rafiee, S.~Dlay, W.~Woo, A review of content-based image retrieval, in: 7th
  International Symposium on Communication Systems, Networks \& Digital Signal
  Processing, 2010, pp. 775--779.
\newblock \href
  {https://doi.org/https://doi.org/10.1109/CSNDSP16145.2010.5580313}
  {\path{doi:https://doi.org/10.1109/CSNDSP16145.2010.5580313}}.

\bibitem{10.1007/978-3-030-32239-7_61}
Y.~Zheng, B.~Jiang, J.~Shi, H.~Zhang, F.~Xie, Encoding histopathological wsis
  using gnn for scalable diagnostically relevant regions retrieval, in:
  D.~Shen, T.~Liu, T.~M. Peters, L.~H. Staib, C.~Essert, S.~Zhou, P.-T. Yap,
  A.~Khan (Eds.), Medical Image Computing and Computer Assisted Intervention --
  MICCAI 2019, Springer International Publishing, Cham, 2019, pp. 550--558.
\newblock \href {https://doi.org/https://doi.org/10.1007/978-3-030-32239-7_61}
  {\path{doi:https://doi.org/10.1007/978-3-030-32239-7_61}}.

\bibitem{tang2020clinician}
S.~Tang, A.~Modi, M.~Sjoding, J.~Wiens, Clinician-in-the-loop decision making:
  Reinforcement learning with near-optimal set-valued policies, in:
  International Conference on Machine Learning, PMLR, 2020, pp. 9387--9396.

\bibitem{Herington1848}
J.~Herington, M.~D. McCradden, K.~Creel, R.~Boellaard, E.~C. Jones, A.~K. Jha,
  A.~Rahmim, P.~J. Scott, J.~J. Sunderland, R.~L. Wahl, S.~Zuehlsdorff,
  B.~Saboury, Ethical considerations for artificial intelligence in medical
  imaging: Data collection, development, and evaluation, Journal of Nuclear
  Medicine 64~(12) (2023) 1848--1854.
\newblock \href {https://doi.org/https://doi.org/10.2967/jnumed.123.266080}
  {\path{doi:https://doi.org/10.2967/jnumed.123.266080}}.

\bibitem{doi:10.1148/rg.253045071}
H.~Müller, A.~Rosset, A.~Garcia, J.-P. Vallée, A.~Geissbuhler, Benefits of
  content-based visual data access in radiology, RadioGraphics 25~(3) (2005)
  849--858.
\newblock \href {https://doi.org/https://doi.org/10.1148/rg.253045071}
  {\path{doi:https://doi.org/10.1148/rg.253045071}}.

\bibitem{9089643}
A.~Belhi, A.~Bouras, {CNN} features vs classical features for largescale
  cultural image retrieval, in: IEEE International Conference on Informatics,
  IoT, and Enabling Technologies, 2020, pp. 95--99.
\newblock \href
  {https://doi.org/https://doi.org/10.1109/ICIoT48696.2020.9089643}
  {\path{doi:https://doi.org/10.1109/ICIoT48696.2020.9089643}}.

\bibitem{10.1145/2647868.2654948}
J.~Wan, D.~Wang, S.~C.~H. Hoi, P.~Wu, J.~Zhu, Y.~Zhang, J.~Li, Deep learning
  for content-based image retrieval: A comprehensive study, in: Proceedings of
  the 22nd ACM International Conference on Multimedia, Association for
  Computing Machinery, 2014, p. 157–166.
\newblock \href {https://doi.org/https://doi.org/10.1145/2647868.2654948}
  {\path{doi:https://doi.org/10.1145/2647868.2654948}}.

\bibitem{9933854}
W.~Chen, Y.~Liu, W.~Wang, E.~M. Bakker, T.~Georgiou, P.~Fieguth, L.~Liu, M.~S.
  Lew, Deep learning for instance retrieval: A survey, IEEE Transactions on
  Pattern Analysis and Machine Intelligence 45~(6) (2023) 7270--7292.
\newblock \href {https://doi.org/https://doi.org/10.1109/TPAMI.2022.3218591}
  {\path{doi:https://doi.org/10.1109/TPAMI.2022.3218591}}.

\bibitem{Kalra2020}
S.~Kalra, H.~R. Tizhoosh, S.~Shah, C.~Choi, S.~Damaskinos, A.~Safarpoor,
  S.~Shafiei, M.~Babaie, P.~Diamandis, C.~J.~V. Campbell, L.~Pantanowitz,
  Pan-cancer diagnostic consensus through searching archival histopathology
  images using artificial intelligence, npj Digital Medicine 3~(1) (2020) 31.
\newblock \href {https://doi.org/https://doi.org/10.1038/s41746-020-0238-2}
  {\path{doi:https://doi.org/10.1038/s41746-020-0238-2}}.

\bibitem{saeidi2024breast}
N.~Saeidi, H.~Karshenas, B.~Shoushtarian, S.~Hatamikia, R.~Woitek, A.~Mahbod,
  Breast histopathology image retrieval by attention-based adversarially
  regularized variational graph autoencoder with contrastive learning-based
  feature extraction, arXiv preprint arXiv:2405.04211 (2024).

\bibitem{denner2024leveraging}
S.~Denner, D.~Zimmerer, D.~Bounias, M.~Bujotzek, S.~Xiao, L.~Kausch,
  P.~Schader, T.~Penzkofer, P.~F. J{\"a}ger, K.~Maier-Hein, Leveraging
  foundation models for content-based medical image retrieval in radiology,
  arXiv preprint arXiv:2403.06567 (2024).

\bibitem{10635170}
F.~K. Jush, T.~Truong, S.~Vogler, M.~Lenga, Medical image retrieval using
  pretrained embeddings, in: IEEE International Symposium on Biomedical
  Imaging, 2024, pp. 1--5.
\newblock \href
  {https://doi.org/https://doi.org/10.1109/ISBI56570.2024.10635170}
  {\path{doi:https://doi.org/10.1109/ISBI56570.2024.10635170}}.

\bibitem{Hegde2019}
N.~Hegde, J.~D. Hipp, Y.~Liu, M.~Emmert-Buck, E.~Reif, D.~Smilkov, M.~Terry,
  C.~J. Cai, M.~B. Amin, C.~H. Mermel, P.~Q. Nelson, L.~H. Peng, G.~S. Corrado,
  M.~C. Stumpe, Similar image search for histopathology: {SMILY}, npj Digital
  Medicine 2~(1) (2019).
\newblock \href {https://doi.org/https://doi.org/10.1038/s41746-019-0131-z}
  {\path{doi:https://doi.org/10.1038/s41746-019-0131-z}}.

\bibitem{Deng2009}
J.~Deng, W.~Dong, R.~Socher, L.-J. Li, K.~Li, L.~Fei-Fei, {ImageNet}: A
  large-scale hierarchical image database, in: IEEE Conference on Computer
  Vision and Pattern Recognition, 2009, pp. 248--255.
\newblock \href {https://doi.org/https://doi.org/10.1109/CVPR.2009.5206848}
  {\path{doi:https://doi.org/10.1109/CVPR.2009.5206848}}.

\bibitem{pmlr-v139-radford21a}
A.~Radford, J.~W. Kim, C.~Hallacy, A.~Ramesh, G.~Goh, S.~Agarwal, G.~Sastry,
  A.~Askell, P.~Mishkin, J.~Clark, G.~Krueger, I.~Sutskever, Learning
  transferable visual models from natural language supervision, in: M.~Meila,
  T.~Zhang (Eds.), Proceedings of the 38th International Conference on Machine
  Learning, Vol. 139 of roceedings of Machine Learning Research, PMLR, 2021,
  pp. 8748--8763.

\bibitem{10692575}
Y.~Qiao, K.~Li, J.~Lin, R.~Wei, C.~Jiang, Y.~Luo, H.~Yang, Robust domain
  generalization for multi-modal object recognition, in: International
  Conference on Artificial Intelligence and Electromechanical Automation, 2024,
  pp. 392--397.
\newblock \href
  {https://doi.org/https://doi.org/10.1109/AIEA62095.2024.10692575}
  {\path{doi:https://doi.org/10.1109/AIEA62095.2024.10692575}}.

\bibitem{9709990}
M.~Caron, H.~Touvron, I.~Misra, H.~Jegou, J.~Mairal, P.~Bojanowski, A.~Joulin,
  Emerging properties in self-supervised vision transformers, in: IEEE/CVF
  International Conference on Computer Vision, 2021, pp. 9630--9640.
\newblock \href {https://doi.org/https://doi.org/10.1109/ICCV48922.2021.00951}
  {\path{doi:https://doi.org/10.1109/ICCV48922.2021.00951}}.

\bibitem{oquab2023dinov2}
M.~Oquab, T.~Darcet, T.~Moutakanni, H.~Vo, M.~Szafraniec, V.~Khalidov,
  P.~Fernandez, D.~Haziza, F.~Massa, A.~El-Nouby, et~al., {DINOv2}: Learning
  robust visual features without supervision, arXiv preprint arXiv:2304.07193
  (2023).

\bibitem{khun2024content}
F.~Khun~Jush, S.~Vogler, T.~Truong, M.~Lenga, Content-based image retrieval for
  multi-class volumetric radiology images: A benchmark study, arXiv e-prints
  (2024) arXiv--2405.

\bibitem{Yang2023}
J.~Yang, R.~Shi, D.~Wei, Z.~Liu, L.~Zhao, B.~Ke, H.~Pfister, B.~Ni, {MedMNIST
  v2} - a large-scale lightweight benchmark for 2d and 3d biomedical image
  classification, Scientific Data 10~(1) (2023) 41.
\newblock \href {https://doi.org/https://doi.org/10.1038/s41597-022-01721-8}
  {\path{doi:https://doi.org/10.1038/s41597-022-01721-8}}.

\bibitem{9434062}
J.~Yang, R.~Shi, B.~Ni, {MedMNIST} classification decathlon: A lightweight
  automl benchmark for medical image analysis, in: IEEE 18th International
  Symposium on Biomedical Imaging, 2021, pp. 191--195.
\newblock \href
  {https://doi.org/https://doi.org/10.1109/ISBI48211.2021.9434062}
  {\path{doi:https://doi.org/10.1109/ISBI48211.2021.9434062}}.

\bibitem{ALDHABYANI2020104863}
W.~Al-Dhabyani, M.~Gomaa, H.~Khaled, A.~Fahmy, Dataset of breast ultrasound
  images, Data in Brief 28 (2020) 104863.
\newblock \href {https://doi.org/https://doi.org/10.1016/j.dib.2019.104863}
  {\path{doi:https://doi.org/10.1016/j.dib.2019.104863}}.

\bibitem{KERMANY20181122}
D.~S. Kermany, M.~Goldbaum, W.~Cai, C.~C. Valentim, H.~Liang, S.~L. Baxter,
  A.~McKeown, G.~Yang, X.~Wu, F.~Yan, J.~Dong, M.~K. Prasadha, J.~Pei, M.~Y.
  Ting, J.~Zhu, C.~Li, S.~Hewett, J.~Dong, I.~Ziyar, A.~Shi, R.~Zhang,
  L.~Zheng, R.~Hou, W.~Shi, X.~Fu, Y.~Duan, V.~A. Huu, C.~Wen, E.~D. Zhang,
  C.~L. Zhang, O.~Li, X.~Wang, M.~A. Singer, X.~Sun, J.~Xu, A.~Tafreshi, M.~A.
  Lewis, H.~Xia, K.~Zhang, Identifying medical diagnoses and treatable diseases
  by image-based deep learning, Cell 172~(5) (2018) 1122--1131.e9.
\newblock \href {https://doi.org/https://doi.org/10.1016/j.cell.2018.02.010}
  {\path{doi:https://doi.org/10.1016/j.cell.2018.02.010}}.

\bibitem{LIU2022100512}
R.~Liu, X.~Wang, Q.~Wu, L.~Dai, X.~Fang, T.~Yan, J.~Son, S.~Tang, J.~Li,
  Z.~Gao, A.~Galdran, J.~Poorneshwaran, H.~Liu, J.~Wang, Y.~Chen, P.~Porwal,
  G.~S. {Wei Tan}, X.~Yang, C.~Dai, H.~Song, M.~Chen, H.~Li, W.~Jia, D.~Shen,
  B.~Sheng, P.~Zhang, {DeepDRiD}: Diabetic retinopathy—grading and image
  quality estimation challenge, Patterns 3~(6) (2022) 100512.
\newblock \href {https://doi.org/https://doi.org/10.1016/j.patter.2022.100512}
  {\path{doi:https://doi.org/10.1016/j.patter.2022.100512}}.

\bibitem{tschandl2018ham10000}
P.~Tschandl, C.~Rosendahl, H.~Kittler, The {HAM10000} dataset, a large
  collection of multi-source dermatoscopic images of common pigmented skin
  lesions, Scientific Data 5 (2018) 180161.
\newblock \href {https://doi.org/https://doi.org/10.1038/sdata.2018.161}
  {\path{doi:https://doi.org/10.1038/sdata.2018.161}}.

\bibitem{ACEVEDO2020105474}
A.~Acevedo, A.~Merino, S.~Alférez, Ángel Molina, L.~Boldú, J.~Rodellar, A
  dataset of microscopic peripheral blood cell images for development of
  automatic recognition systems, Data in Brief 30 (2020) 105474.
\newblock \href {https://doi.org/https://doi.org/10.1016/j.dib.2020.105474}
  {\path{doi:https://doi.org/10.1016/j.dib.2020.105474}}.

\bibitem{10.1371/journal.pmed.1002730}
J.~N. Kather, J.~Krisam, P.~Charoentong, T.~Luedde, E.~Herpel, C.-A. Weis,
  T.~Gaiser, A.~Marx, N.~A. Valous, D.~Ferber, L.~Jansen, C.~C. Reyes-Aldasoro,
  I.~Zörnig, D.~Jäger, H.~Brenner, J.~Chang-Claude, M.~Hoffmeister,
  N.~Halama, Predicting survival from colorectal cancer histology slides using
  deep learning: A retrospective multicenter study, PLOS Medicine 16~(1) (2019)
  1--22.
\newblock \href {https://doi.org/https://doi.org/10.1371/journal.pmed.1002730}
  {\path{doi:https://doi.org/10.1371/journal.pmed.1002730}}.

\bibitem{Simonyan2014}
K.~Simonyan, A.~Zisserman, Very deep convolutional networks for large-scale
  image recognition, arXiv preprint arXiv:1409.1556 (2014).

\bibitem{He2016}
K.~He, X.~Zhang, S.~Ren, J.~Sun, Deep residual learning for image recognition,
  in: IEEE Conference on Computer Vision and Pattern Recognition, 2016, pp.
  770--778.
\newblock \href {https://doi.org/https://doi.org/10.1109/CVPR.2016.90}
  {\path{doi:https://doi.org/10.1109/CVPR.2016.90}}.

\bibitem{huang2017densely}
G.~Huang, Z.~Liu, L.~Van Der~Maaten, K.~Q. Weinberger, Densely connected
  convolutional networks, in: IEEE Conference on Computer Vision and Pattern
  Recognition, 2017, pp. 2261--2269.
\newblock \href {https://doi.org/https://doi.org/10.1109/CVPR.2017.243}
  {\path{doi:https://doi.org/10.1109/CVPR.2017.243}}.

\bibitem{Tan2021}
M.~Tan, Q.~Le, {EfficientNetV2}: Smaller models and faster training, in:
  M.~Meila, T.~Zhang (Eds.), Proceedings of the 38th International Conference
  on Machine Learning, Vol. 139 of Proceedings of Machine Learning Research,
  PMLR, 2021, pp. 10096--10106.

\bibitem{wang2022medclip}
Z.~Wang, Z.~Wu, D.~Agarwal, J.~Sun, {MedCLIP}: Contrastive learning from
  unpaired medical images and text, arXiv preprint arXiv:2210.10163 (2022).

\bibitem{zhang2023large}
S.~Zhang, Y.~Xu, N.~Usuyama, H.~Xu, J.~Bagga, R.~Tinn, S.~Preston, R.~Rao,
  M.~Wei, N.~Valluri, et~al., {BiomedCLIP}: a multimodal biomedical foundation
  model pretrained from fifteen million scientific image-text pairs, arXiv
  preprint arXiv:2303.00915 (2023).

\bibitem{ilharco_gabriel_2021_5143773}
G.~Ilharco, M.~Wortsman, R.~Wightman, C.~Gordon, N.~Carlini, R.~Taori, A.~Dave,
  V.~Shankar, H.~Namkoong, J.~Miller, H.~Hajishirzi, A.~Farhadi, L.~Schmidt,
  Openclip (Jul. 2021).
\newblock \href {https://doi.org/https://doi.org/10.5281/zenodo.5143773}
  {\path{doi:https://doi.org/10.5281/zenodo.5143773}}.

\bibitem{10205297}
M.~Cherti, R.~Beaumont, R.~Wightman, M.~Wortsman, G.~Ilharco, C.~Gordon,
  C.~Schuhmann, L.~Schmidt, J.~Jitsev, Reproducible scaling laws for
  contrastive language-image learning, in: IEEE/CVF Conference on Computer
  Vision and Pattern Recognition, 2023, pp. 2818--2829.
\newblock \href {https://doi.org/10.1109/CVPR52729.2023.00276}
  {\path{doi:10.1109/CVPR52729.2023.00276}}.

\bibitem{Lu2024}
M.~Y. Lu, B.~Chen, D.~F.~K. Williamson, R.~J. Chen, I.~Liang, T.~Ding,
  G.~Jaume, I.~Odintsov, L.~P. Le, G.~Gerber, A.~V. Parwani, A.~Zhang,
  F.~Mahmood, A visual-language foundation model for computational pathology,
  Nature Medicine 30~(3) (2024) 863--874.
\newblock \href {https://doi.org/https://doi.org/10.1038/s41591-024-02856-4}
  {\path{doi:https://doi.org/10.1038/s41591-024-02856-4}}.

\bibitem{Chen2024}
R.~J. Chen, T.~Ding, M.~Y. Lu, D.~F.~K. Williamson, G.~Jaume, A.~H. Song,
  B.~Chen, A.~Zhang, D.~Shao, M.~Shaban, M.~Williams, L.~Oldenburg, L.~L.
  Weishaupt, J.~J. Wang, A.~Vaidya, L.~P. Le, G.~Gerber, S.~Sahai, W.~Williams,
  F.~Mahmood, Towards a general-purpose foundation model for computational
  pathology, Nature Medicine 30~(3) (2024) 850--862.
\newblock \href {https://doi.org/https://doi.org/10.1038/s41591-024-02857-3}
  {\path{doi:https://doi.org/10.1038/s41591-024-02857-3}}.

\bibitem{Kirillov_2023_ICCV}
A.~Kirillov, E.~Mintun, N.~Ravi, H.~Mao, C.~Rolland, L.~Gustafson, T.~Xiao,
  S.~Whitehead, A.~C. Berg, W.-Y. Lo, P.~Dollar, R.~Girshick, Segment anything,
  in: Proceedings of the IEEE/CVF International Conference on Computer Vision,
  2023, pp. 4015--4026.
\newblock \href {https://doi.org/https://doi.org/10.1109/ICCV51070.2023.00371}
  {\path{doi:https://doi.org/10.1109/ICCV51070.2023.00371}}.

\bibitem{Ma2024}
J.~Ma, Y.~He, F.~Li, L.~Han, C.~You, B.~Wang, Segment anything in medical
  images, Nature Communications 15~(1) (2024) 654.
\newblock \href {https://doi.org/https://doi.org/10.1038/s41467-024-44824-z}
  {\path{doi:https://doi.org/10.1038/s41467-024-44824-z}}.

\bibitem{Vorontsov2024}
E.~Vorontsov, A.~Bozkurt, A.~Casson, G.~Shaikovski, M.~Zelechowski,
  K.~Severson, E.~Zimmermann, J.~Hall, N.~Tenenholtz, N.~Fusi, E.~Yang,
  P.~Mathieu, A.~van Eck, D.~Lee, J.~Viret, E.~Robert, Y.~K. Wang, J.~D. Kunz,
  M.~C.~H. Lee, J.~H. Bernhard, R.~A. Godrich, G.~Oakley, E.~Millar, M.~Hanna,
  H.~Wen, J.~A. Retamero, W.~A. Moye, R.~Yousfi, C.~Kanan, D.~S. Klimstra,
  B.~Rothrock, S.~Liu, T.~J. Fuchs, A foundation model for clinical-grade
  computational pathology and rare cancers detection, Nature Medicine (2024).
\newblock \href {https://doi.org/https://doi.org/10.1038/s41591-024-03141-0}
  {\path{doi:https://doi.org/10.1038/s41591-024-03141-0}}.

\bibitem{9879206}
K.~He, X.~Chen, S.~Xie, Y.~Li, P.~Dollár, R.~Girshick, Masked autoencoders are
  scalable vision learners, in: IEEE/CVF Conference on Computer Vision and
  Pattern Recognition, 2022, pp. 15979--15988.
\newblock \href {https://doi.org/https://doi.org/10.1109/CVPR52688.2022.01553}
  {\path{doi:https://doi.org/10.1109/CVPR52688.2022.01553}}.

\bibitem{dosovitskiy2020image}
A.~Dosovitskiy, L.~Beyer, A.~Kolesnikov, D.~Weissenborn, X.~Zhai,
  T.~Unterthiner, M.~Dehghani, M.~Minderer, G.~Heigold, S.~Gelly, et~al., An
  image is worth 16x16 words: Transformers for image recognition at scale,
  arXiv preprint arXiv:2010.11929 (2020).

\bibitem{9710580}
Z.~Liu, Y.~Lin, Y.~Cao, H.~Hu, Y.~Wei, Z.~Zhang, S.~Lin, B.~Guo, {Swin
  Transformer}: Hierarchical vision transformer using shifted windows, in:
  IEEE/CVF International Conference on Computer Vision, 2021, pp. 9992--10002.
\newblock \href {https://doi.org/https://doi.org/10.1109/ICCV48922.2021.00986}
  {\path{doi:https://doi.org/10.1109/ICCV48922.2021.00986}}.

\bibitem{ALZUBI201795}
A.~Alzu'bi, A.~Amira, N.~Ramzan, Content-based image retrieval with compact
  deep convolutional features, Neurocomputing 249 (2017) 95--105.
\newblock \href {https://doi.org/https://doi.org/10.1016/j.neucom.2017.03.072}
  {\path{doi:https://doi.org/10.1016/j.neucom.2017.03.072}}.

\bibitem{Agrawal2022}
S.~Agrawal, A.~Chowdhary, S.~Agarwala, V.~Mayya, S.~Kamath~S., Content-based
  medical image retrieval system for lung diseases using deep cnns,
  International Journal of Information Technology 14~(7) (2022) 3619--3627.
\newblock \href {https://doi.org/https://doi.org/10.1007/s41870-022-01007-7}
  {\path{doi:https://doi.org/10.1007/s41870-022-01007-7}}.

\bibitem{mahbod2018breast}
A.~Mahbod, I.~Ellinger, R.~Ecker, {\"O}.~Smedby, C.~Wang, Breast cancer
  histological image classification using fine-tuned deep network fusion, in:
  A.~Campilho, F.~Karray, B.~ter Haar~Romeny (Eds.), Image Analysis and
  Recognition, Springer International Publishing, Cham, 2018, pp. 754--762.
\newblock \href {https://doi.org/https://doi.org/10.1007/978-3-319-93000-8_85}
  {\path{doi:https://doi.org/10.1007/978-3-319-93000-8_85}}.

\bibitem{9313211}
A.~Mbilinyi, H.~Schuldt, Cross-modality medical image retrieval with deep
  features, in: IEEE International Conference on Bioinformatics and
  Biomedicine, 2020, pp. 2632--2639.
\newblock \href
  {https://doi.org/https://doi.org/10.1109/BIBM49941.2020.9313211}
  {\path{doi:https://doi.org/10.1109/BIBM49941.2020.9313211}}.

\bibitem{9412307}
A.~Mahbod, G.~Schaefer, C.~Wang, R.~Ecker, G.~Dorffner, I.~Ellinger,
  Investigating and exploiting image resolution for transfer learning-based
  skin lesion classification, in: International Conference on Pattern
  Recognition, 2021, pp. 4047--4053.
\newblock \href
  {https://doi.org/https://doi.org/10.1109/ICPR48806.2021.9412307}
  {\path{doi:https://doi.org/10.1109/ICPR48806.2021.9412307}}.

\bibitem{tan2019efficientnet}
M.~Tan, Q.~Le, {EfficientNet}: Rethinking model scaling for convolutional
  neural networks, in: K.~Chaudhuri, R.~Salakhutdinov (Eds.), Proceedings of
  the 36th International Conference on Machine Learning, Vol.~97 of Proceedings
  of Machine Learning Research, PMLR, 2019, pp. 6105--6114.

\bibitem{zoph2016neural}
B.~Zoph, Q.~V. Le, Neural architecture search with reinforcement learning,
  arXiv preprint arXiv:1611.01578 (2016).

\bibitem{Ekoputris2018}
M.~Sandler, A.~Howard, M.~Zhu, A.~Zhmoginov, L.-C. Chen, {MobileNetV2}:
  Inverted residuals and linear bottlenecks, in: Proceedings of the IEEE
  Conference on Computer Vision and Pattern Recognition, 2018.
\newblock \href {https://doi.org/https://doi.org/10.1109/CVPR.2018.00474}
  {\path{doi:https://doi.org/10.1109/CVPR.2018.00474}}.

\bibitem{gupta2020accelerator}
S.~Gupta, B.~Akin, Accelerator-aware neural network design using automl, arXiv
  preprint arXiv:2003.02838 (2020).

\bibitem{10.1007/978-3-031-20233-9_45}
Y.~Xin, Y.~Tang, Z.~Yang, Shoe print retrieval algorithm based on improved
  {EfficientnetV2}, in: W.~Deng, J.~Feng, D.~Huang, M.~Kan, Z.~Sun, F.~Zheng,
  W.~Wang, Z.~He (Eds.), Biometric Recognition, Springer Nature Switzerland,
  Cham, 2022, pp. 444--454.
\newblock \href {https://doi.org/https://doi.org/10.1007/978-3-031-20233-9_45}
  {\path{doi:https://doi.org/10.1007/978-3-031-20233-9_45}}.

\bibitem{10.1136/jamia.2009.002733}
A.~R. Aronson, F.-M. Lang, An overview of {MetaMap}: historical perspective and
  recent advances, Journal of the American Medical Informatics Association
  17~(3) (2010) 229--236.
\newblock \href {https://doi.org/https://doi.org/10.1136/jamia.2009.002733}
  {\path{doi:https://doi.org/10.1136/jamia.2009.002733}}.

\bibitem{10.1145/3458754}
Y.~Gu, R.~Tinn, H.~Cheng, M.~Lucas, N.~Usuyama, X.~Liu, T.~Naumann, J.~Gao,
  H.~Poon, Domain-specific language model pretraining for biomedical natural
  language processing, ACM Trans. Comput. Healthcare 3~(1) (2021).
\newblock \href {https://doi.org/https://doi.org/10.1145/3458754}
  {\path{doi:https://doi.org/10.1145/3458754}}.

\bibitem{schuhmann2022laionb}
C.~Schuhmann, R.~Beaumont, R.~Vencu, C.~W. Gordon, R.~Wightman, M.~Cherti,
  T.~Coombes, A.~Katta, C.~Mullis, M.~Wortsman, P.~Schramowski, S.~R.
  Kundurthy, K.~Crowson, L.~Schmidt, R.~Kaczmarczyk, J.~Jitsev,
  \href{https://openreview.net/forum?id=M3Y74vmsMcY}{{LAION}-5b: An open
  large-scale dataset for training next generation image-text models}, in:
  Thirty-sixth Conference on Neural Information Processing Systems Datasets and
  Benchmarks Track, 2022.
\newline\urlprefix\url{https://openreview.net/forum?id=M3Y74vmsMcY}

\bibitem{9880094}
X.~Zhai, A.~Kolesnikov, N.~Houlsby, L.~Beyer, Scaling vision transformers, in:
  IEEE/CVF Conference on Computer Vision and Pattern Recognition, 2022, pp.
  1204--1213.
\newblock \href {https://doi.org/https://doi.org/10.1109/CVPR52688.2022.01179}
  {\path{doi:https://doi.org/10.1109/CVPR52688.2022.01179}}.

\bibitem{7780460}
J.~Redmon, S.~Divvala, R.~Girshick, A.~Farhadi, You only look once: Unified,
  real-time object detection, in: IEEE Conference on Computer Vision and
  Pattern Recognition, 2016, pp. 779--788.
\newblock \href {https://doi.org/https://doi.org/10.1109/CVPR.2016.91}
  {\path{doi:https://doi.org/10.1109/CVPR.2016.91}}.

\bibitem{10.1093/bib/bbac409}
R.~Luo, L.~Sun, Y.~Xia, T.~Qin, S.~Zhang, H.~Poon, T.-Y. Liu, {BioGPT}:
  generative pre-trained transformer for biomedical text generation and mining,
  Briefings in Bioinformatics 23~(6) (2022) bbac409.
\newblock \href {https://doi.org/https://doi.org/10.1093/bib/bbac409}
  {\path{doi:https://doi.org/10.1093/bib/bbac409}}.

\bibitem{yu2022coca}
J.~Yu, Z.~Wang, V.~Vasudevan, L.~Yeung, M.~Seyedhosseini, Y.~Wu, {CoCa}:
  Contrastive captioners are image-text foundation models, arXiv preprint
  arXiv:2205.01917 (2022).

\bibitem{zhou2021ibot}
J.~Zhou, C.~Wei, H.~Wang, W.~Shen, C.~Xie, A.~Yuille, T.~Kong, {iBOT}: Image
  bert pre-training with online tokenizer, arXiv preprint arXiv:2111.07832
  (2021).

\bibitem{MACKIEWICZ1993303}
A.~Maćkiewicz, W.~Ratajczak, Principal components analysis (pca), Computers \&
  Geosciences 19~(3) (1993) 303--342.
\newblock \href {https://doi.org/https://doi.org/10.1016/0098-3004(93)90090-R}
  {\path{doi:https://doi.org/10.1016/0098-3004(93)90090-R}}.

\bibitem{WANG2016232}
Y.~Wang, H.~Yao, S.~Zhao, Auto-encoder based dimensionality reduction,
  Neurocomputing 184 (2016) 232--242.
\newblock \href {https://doi.org/https://doi.org/10.1016/j.neucom.2015.08.104}
  {\path{doi:https://doi.org/10.1016/j.neucom.2015.08.104}}.

\bibitem{JMLR:v9:vandermaaten08a}
L.~van~der Maaten, G.~Hinton, Visualizing data using {t-SNE}, Journal of
  Machine Learning Research 9~(86) (2008) 2579--2605.

\bibitem{McInnes2018}
L.~McInnes, J.~Healy, N.~Saul, L.~Großberger, {UMAP}: Uniform manifold
  approximation and projection, Journal of Open Source Software 3~(29) (2018)
  861.
\newblock \href {https://doi.org/https://doi.org/10.21105/joss.00861}
  {\path{doi:https://doi.org/10.21105/joss.00861}}.

\bibitem{WANG2023102645}
X.~Wang, Y.~Du, S.~Yang, J.~Zhang, M.~Wang, J.~Zhang, W.~Yang, J.~Huang,
  X.~Han, {RetCCL}: Clustering-guided contrastive learning for whole-slide
  image retrieval, Medical Image Analysis 83 (2023) 102645.
\newblock \href {https://doi.org/https://doi.org/10.1016/j.media.2022.102645}
  {\path{doi:https://doi.org/10.1016/j.media.2022.102645}}.

\bibitem{mahbod2021pollen}
A.~Mahbod, G.~Schaefer, R.~Ecker, I.~Ellinger, Pollen grain microscopic image
  classification using an ensemble of fine-tuned deep convolutional neural
  networks, in: International Conference on Pattern Recognition, Springer,
  2021, pp. 344--356.
\newblock \href {https://doi.org/https://doi.org/10.1007/978-3-030-68763-2_26}
  {\path{doi:https://doi.org/10.1007/978-3-030-68763-2_26}}.

\bibitem{MAHBOD2020105475}
A.~Mahbod, G.~Schaefer, C.~Wang, G.~Dorffner, R.~Ecker, I.~Ellinger, Transfer
  learning using a multi-scale and multi-network ensemble for skin lesion
  classification, Computer Methods and Programs in Biomedicine 193 (2020)
  105475.
\newblock \href {https://doi.org/https://doi.org/10.1016/j.cmpb.2020.105475}
  {\path{doi:https://doi.org/10.1016/j.cmpb.2020.105475}}.

\bibitem{XIE2024103023}
Y.~Xie, J.~Zhang, L.~Liu, H.~Wang, Y.~Ye, J.~Verjans, Y.~Xia, {ReFs}: A hybrid
  pre-training paradigm for 3d medical image segmentation, Medical Image
  Analysis 91 (2024) 103023.
\newblock \href {https://doi.org/https://doi.org/10.1016/j.media.2023.103023}
  {\path{doi:https://doi.org/10.1016/j.media.2023.103023}}.

\bibitem{10242007}
Y.~Xie, J.~Zhang, Y.~Xia, C.~Shen, Learning from partially labeled data for
  multi-organ and tumor segmentation, IEEE Transactions on Pattern Analysis and
  Machine Intelligence 45~(12) (2023) 14905--14919.
\newblock \href {https://doi.org/https://doi.org/10.1109/TPAMI.2023.3312587}
  {\path{doi:https://doi.org/10.1109/TPAMI.2023.3312587}}.

\bibitem{10605840}
Y.~Ye, J.~Zhang, Z.~Chen, Y.~Xia, {CADS}: A self-supervised learner via
  cross-modal alignment and deep self-distillation for ct volume segmentation,
  IEEE Transactions on Medical Imaging 44~(1) (2025) 118--129.
\newblock \href {https://doi.org/https://doi.org/10.1109/TMI.2024.3431916}
  {\path{doi:https://doi.org/10.1109/TMI.2024.3431916}}.

\bibitem{10617802}
Y.~Xie, J.~Zhang, Y.~Xia, Q.~Wu, {UniMiSS+}: Universal medical self-supervised
  learning from cross-dimensional unpaired data, IEEE Transactions on Pattern
  Analysis and Machine Intelligence 46~(12) (2024) 10021--10035.
\newblock \href {https://doi.org/https://doi.org/10.1109/TPAMI.2024.3436105}
  {\path{doi:https://doi.org/10.1109/TPAMI.2024.3436105}}.

\bibitem{jush2024content}
F.~K. Jush, S.~Vogler, T.~Truong, M.~Lenga, Content-based image retrieval for
  multi-class volumetric radiology images: A benchmark study, arXiv preprint
  arXiv:2405.09334 (2024).

\bibitem{isensee2021nnu}
F.~Isensee, P.~F. Jaeger, S.~A.~A. Kohl, J.~Petersen, K.~H. Maier-Hein,
  nnu-net: a self-configuring method for deep learning-based biomedical image
  segmentation, Nature Methods 18~(2) (2021) 203--211.
\newblock \href {https://doi.org/https://doi.org/10.1038/s41592-020-01008-z}
  {\path{doi:https://doi.org/10.1038/s41592-020-01008-z}}.

\end{thebibliography}

\end{document}